\documentclass[letterpaper,journal]{IEEEtran}

\usepackage{amsmath,amsfonts}
\usepackage{algorithm}
\usepackage{array}
\usepackage[caption=false,font=normalsize,labelfont=sf,textfont=sf]{subfig}
\usepackage{textcomp}
\usepackage{stfloats}
\usepackage{url}
\usepackage{verbatim}
\usepackage{graphicx}
\usepackage{cite}
\usepackage{url, hyperref, cleveref}
\hypersetup{
    colorlinks=true,
    linkcolor=blue,
    filecolor=magenta,      
    urlcolor=blue,
    citecolor=blue,
}
\usepackage{booktabs}
\usepackage[dvipsnames]{xcolor}
\usepackage{float}
\usepackage{enumitem}
\usepackage[noend]{algpseudocode}
\usepackage{amsmath}
\usepackage{siunitx}
\usepackage{multirow}
\usepackage{comment}
\usepackage{fancyhdr}

\algrenewcommand{\algorithmiccomment}[1]{\hfill\colorbox{light-gray}{$\triangleright$ #1}}

\newcommand\new[1]{\textcolor{Black}{#1}}  

\definecolor{light-gray}{gray}{0.92}
\definecolor{dark-gray}{gray}{0.85}
\definecolor{blue}{HTML}{2070b4}
\definecolor{red}{HTML}{ca171c}

\begin{document}

\title{Adversarial Coevolutionary Illumination with\\ Generational Adversarial MAP-Elites}


\author{\IEEEauthorblockN{
    Timothée Anne$^{1}$,
    Noah Syrkis$^{1}$,
    Meriem Elhosni$^{2}$,
    Florian Turati$^{2}$, \\
    {Franck Legendre$^{2}$, 
    Alain Jaquier$^{2}$, \and
    Sebastian Risi$^{1}$ }\\
    \mbox{}\\
    $^{1}$IT University of Copenhagen, Copenhagen, Denmark \\
    $^{2}$armasuisse Science+Technology, Thun, Switzerland \\
    Corresponding author: sebr@itu.dk
} }



\maketitle

\begin{abstract}
Quality-Diversity (QD) algorithms seek to discover diverse, high-performing solutions across a behavior space, in contrast to conventional optimization methods that target a single optimum. Adversarial problems present unique challenges for QD approaches, as the competing nature of opposing sides creates interdependencies that complicate the evolution process. Existing QD methods applied to such scenarios typically fix one side, constraining the open-endedness. We present Generational Adversarial MAP-Elites (GAME), a coevolutionary QD algorithm that evolves both sides by alternating which side is evolved at each generation. By integrating a vision embedding model (VEM), our approach eliminates the need for domain-specific behavior descriptors and instead operates on video. We validate GAME across three distinct adversarial domains: a multi-agent battle game, a soft-robot wrestling environment, and a deck building game. 
\new{We validate that all its components are necessary, that the VEM is effective in two different domains, and that GAME finds better solutions than one-sided QD baselines.} Our experiments reveal several evolutionary phenomena, including arms race-like dynamics, enhanced novelty through generational extinction, and the preservation of neutral mutations as crucial stepping stones toward the highest performance. While GAME successfully illuminates all three adversarial problems, its capacity for truly open-ended discovery remains constrained by the nature of the search spaces used in this paper. These findings show GAME's broad applicability and highlight opportunities for future research into open-ended adversarial coevolution. Code and videos are available at: \url{https://github.com/Timothee-ANNE/GAME}
\end{abstract}


\section{Introduction}

Evolving a set of diverse, high-quality solutions that optimize a fitness function while covering a behavior space is known as Quality-Diversity (QD)~\cite{pugh2016quality} (or Illumination). It has been successfully applied in multiple domains: robotics~\cite{cully2015robots}, video games~\cite{gravina2019procedural}, chemical synthesis~\cite{jiang2022artificial}, and aeronautics~\cite{brevault2024bayesian}.

Only a few works have applied QD to adversarial problems, even though they are common in multiple domains: military conflicts~\cite{schelling1980strategy}, economics~\cite{baldwin2011understanding}, deep learning~\cite{chakraborty2018adversarial}, or cybersecurity~\cite{li2021arms}.
For example, in the card battle game Hearthstone~\cite{hearthstone2014}, \cite{fontaine2019mapping} evolve decks, while \cite{fontaine2020covariance} evolve neural network policies to play fixed decks. More recently, QD has been used to search for malicious prompts against a Large Language Model (LLM)~\cite{Samvelyan2024RainbowTOA}. In addition, those approaches always fix one side of the adversarial problem. This limits the breadth of illumination, as adversarial domains are often subject to arms races~\cite{dawkins1979arms}, requiring both sides to adapt.

Evolving both sides follows the artificial life goal of creating an open-ended evolutionary process through adversarial coevolution~\cite{bedau2000open, dorin2024artificial}. For example, the POET algorithm~\cite{wang2019poet, wang2020enhanced} coevolves bipedal walkers' policies and environments to create an automatic curriculum toward more robust policies; \cite{costa2020exploring} coevolve Generative Adversarial Networks (GANs) to improve their performance using QD; and recently, \cite{dharna2024quality} coevolve Python scripts in a pursuer and evader game. However, a limitation of those works is that they are domain-specific.

\begin{figure}[t]
    \centering
    \includegraphics[width=0.8\linewidth]{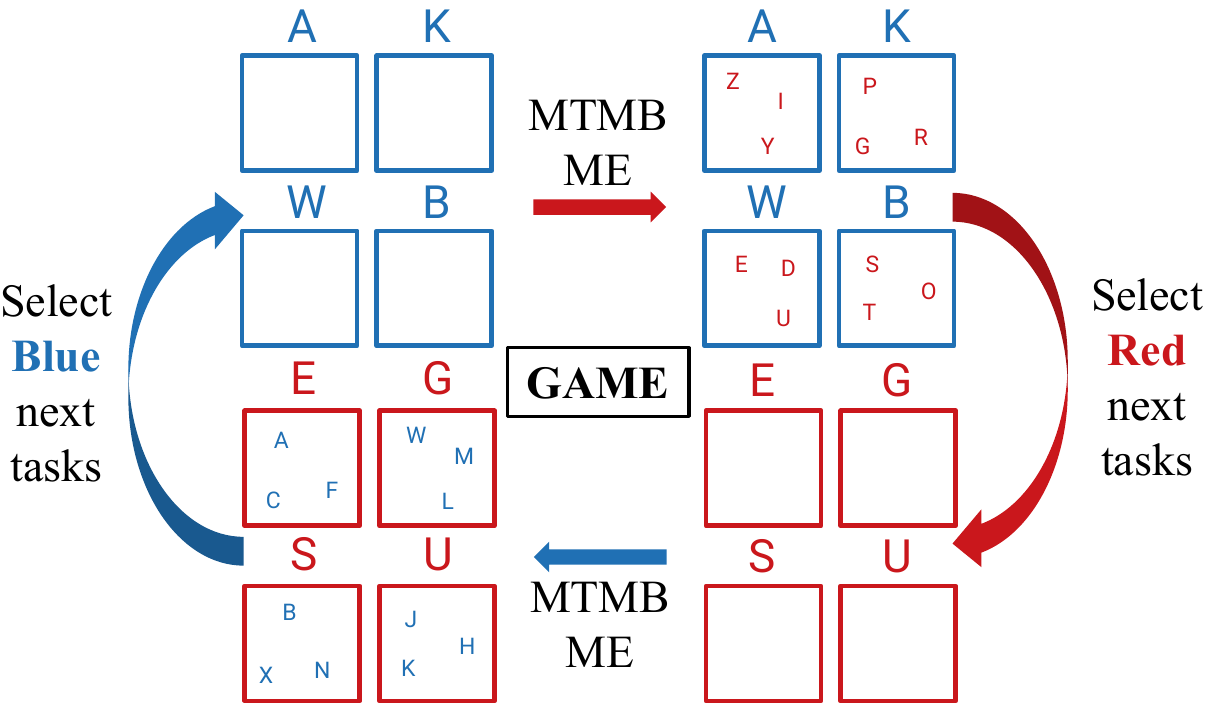}
    \vspace{-2mm}
    \caption{\textbf{GAME's core idea} is the iterative illumination of an adversarial problem using MTMB-ME~\cite{anne2023multi}, switching sides at each generation to promote arms race dynamics. 
    Here, each letter corresponds to a different solution.}
    \label{fig:GAME_core_idea}
   \vspace{-5mm}
\end{figure}

\new{We present \emph{Generational Adversarial MAP-Elites} (GAME), a new coevolutionary QD algorithm that illuminates both sides of an adversarial problem by alternating the evolution of solutions on one side that maximize the adversarial fitness against fixed opponents from the other side (Fig.~\ref{fig:GAME_core_idea}). A second contribution is the use of a Vision Embedding Model (VEM) (we use CLIP~\cite{radford2021learning}) to obtain a domain-agnostic behavior descriptor from videos, without requiring a handcrafted behavior descriptor. To handle the high dimensionality of the embedding space, GAME uses an unstructured archive~\cite{vassiliades2017comparison}.}

\new{This paper extends the conference paper~\cite{anne2025adversarial}, which introduced GAME. In addition to the original experiments on a multi-agent battle game called Parabellum~\cite{anne2025harnessing}, we evaluate GAME on two additional adversarial problems: (1) a 2D soft-robot custom EvoGym~\cite{bhatia2021evolution} environment called Wrestling and (2) a deck building game called Hearthbreaker~\cite{hearthbreaker}, which is a Python simulator of Hearthstone~\cite{hearthstone2014}.}
 
\new{The contributions are: (1) the original experiments in Parabellum, which demonstrate that all components of GAME are necessary to illuminate the space of possible strategies, (2) illumination takeaways, and (3) an example of an arms race dynamic. In addition, this paper's new contributions are: (4) a comparison in Parabellum with a second baseline that does not use the VEM, further validating the VEM's usefulness; in Wrestling, (5) GAME's promise for the evolution of artificial creatures; (6) validation of the use of an unstructured archive; and (7) the assessment of the VEM behavior's robustness to the frame rate used; and (8) a comparison with a one-sided illumination in Wrestling and Hearthbreaker, showing that GAME yields higher-quality solutions but lower coverage.}
    
\section{Related Work}

\subsection{Artificial Life, Open-endedness, and Coevolution}
Creating an open-ended process that continually presents new, interesting challenges and, in turn, fosters the evolution of novel, meaningful solutions to those challenges is a central goal in the artificial life community~\cite{bedau2000open, dorin2024artificial}.

One way to move toward open-endedness is through coevolution driven by arms race dynamics~\cite{dawkins1979arms}. A key difficulty lies in finding a balance that avoids the collapse of one side or the emergence of a stable equilibrium where both sides cease to innovate~\cite{ficici1998challenges}.
\cite{moran2019evolving} demonstrate that a blend of cooperation and competition can lead to greater complexity growth. Similarly, \cite{harrington2019escalation} show that greater policy evolvability supports a longer-lasting arms race and increased complexity. 

A related but not strictly adversarial setting is the coevolution of an agent and its environment. The objective is for the environment to evolve in a way that presents new challenges that are neither too easy nor too difficult, effectively creating a curriculum for the agent by offering a sequence of meaningful stepping stones.
\cite{brant2017minimal} coevolve mazes and maze-solving agents using minimal criterion coevolution, demonstrating that no handcrafted fitness function or behavior descriptor is required.

Building on this idea, POET~\cite{wang2019poet} generates an open-ended sequence of environments for training bipedal walkers. It maintains a population of active environment–controller pairs. New environments are generated from recently active ones that have shown sufficient progress, where the task is neither easy nor difficult for the associated controller. Each controller is independently optimized using an evolutionary strategy. POET also facilitates transfer between pairs, attempting to apply controllers trained in one environment to others.

Enhanced-POET~\cite{wang2020enhanced} introduces a new measure of novelty for environments, called Performance of All Transferred Agents – Environment Comparison (PATA-EC). This metric requires only fitness scores, eliminating the need for handcrafted behavior descriptors. It is based on a round-robin tournament of agents, comparing their performance rankings across environments. The underlying idea is that a novel environment should yield a different ranking.  

\new{More recently, \cite{chigot2022coevolution} coevolve agents and environments defined by neural cellular automata and show that coevolution leads to more robust and general solutions than direct evolution in challenging environments. OMNI-EPIC~\cite{faldoromni} uses LLMs to generate environments and rewards, thereby creating an open-ended curriculum for reinforcement learning agents. It updates an archive of environments, which helps an LLM estimate the interestingness of a task (i.e., novel and balanced in difficulty).}

\new{These works focus on agent–environment problems without attempting to illuminate both sides. In contrast, GAME explicitly aims to illuminate both sides of any adversarial problem.}

\subsection{Self-Play}

Self-play is a related field that develops a learning curriculum by having the agent play against itself or earlier versions of itself. AlphaStar~\cite{vinyals2019grandmaster} achieves grandmaster level in the real-time strategy (RTS) player-versus-player game StarCraft 2. It uses a league of agents to prevent strategy collapse, improve robustness, and avoid local minima and forgetting. 

\cite{baker2019emergent} propose an RL framework for adversarial self-play in a 3D multi-agent hide-and-seek game. It enabled the emergence of six strategic phases as both sides found new ways to counter each other's high-performing behaviors. The goal of self-play, however, is to learn a robust, high-performing policy, not to illuminate all possible policies.

\subsection{Quality-Diversity for adversarial problems}
Two popular QD algorithms are Novelty Search with Local Competition (NSLC)~\cite{lehman2011evolving} and Multidimensional Archive of Phenotypic Elites (MAP-Elites)~\cite{mouret2015illuminating}. MAP-Elites works by constructing an archive of high-performing solutions, known as elites, organized into cells that discretize the behavior space. MAP-Elites generates a new candidate solution at each iteration using variation operators, evaluates it, and, if it exhibits novel behavior or has higher fitness than the current elite in its corresponding cell, it becomes that cell's elite.

Multi-Task MAP-Elites (MT-ME)~\cite{mouret2020quality} addresses multi-task problems, i.e., the simultaneous optimization of multiple fitness functions. The idea is that solving these tasks simultaneously can improve sampling efficiency, as related tasks may share similar solutions. Building on this, Multi-task Multi-behavior MAP-Elites (MTMB-ME)~\cite{anne2023multi} extends MT-ME by evolving diverse high-performing solutions for each task. 
In this paper, GAME uses MTMB-ME at each generation to evolve solutions that compete against a fixed set of solutions from the previous generation (i.e., the tasks).

QD algorithms have been used to illuminate adversarial problems. For example, \cite{fontaine2019mapping} propose MAP-Elites with sliding boundaries to build different decks in Hearthstone~\cite{hearthstone2014}, using as opponents first a fixed set of decks and then the best decks found in the first phase to explore the possible counters. Using the six best found decks as opponents, \cite{fontaine2020covariance} use Covariance Matrix Adaptation MAP-Elites to evolve a diverse set of neural network controllers and find different strategies when playing with one deck.
\cite{steckel2021illuminating} use MAP-Elites to illuminate the space for generated levels by different GANs in the Lone Runner game. 
\cite{wan2024quality} apply QD principles to adversarial imitation learning to train agents on several behaviors simultaneously and show that it can potentially improve any inverse RL method. 
\cite{samvelyan2024multi} apply MAP-Elites to explore adversarial scenarios for a pre-trained DRL agent in a multi-agent game environment (Google Research Football), uncovering strategic weaknesses in the agent's behavior.
\cite{Samvelyan2024RainbowTOA} use MT-ME to find a set of diverse adversarial safety attacks on an LLM, using an LLM for the variation operator and for evaluating the fitness of the attack. These methods only optimize one side of the adversarial problem, significantly limiting illumination. \new{Our second and third case studies are initial studies to compare GAME against a one-sided illumination similar to \cite{fontaine2019mapping}. 
} 

Few QD methods attempt to coevolve both sides. \cite{costa2020exploring} use NSLC to coevolve GANs (i.e., the generators and descriptors), showing that it improves diversity and discovers better models. Closer to our work, \cite{dharna2024quality} propose a self-play QD algorithm to coevolve adversarial Python script controllers for an asymmetric adversarial game between pursuer and evader agents. They use the embedding space of an LLM to model genotypic diversity, whereas in this paper we are interested in behavioral diversity. Their work also uses an unstructured archive to handle the high-dimensional embedding space. However, they employ an NSLC-type unstructured archive, which requires more problem-specific tuning of the novelty threshold parameter than the growing archives used in GAME.

\subsection{Using a VEM for behavior space}

\begin{figure*}[ht]
    \centering
    \includegraphics[width=\linewidth]{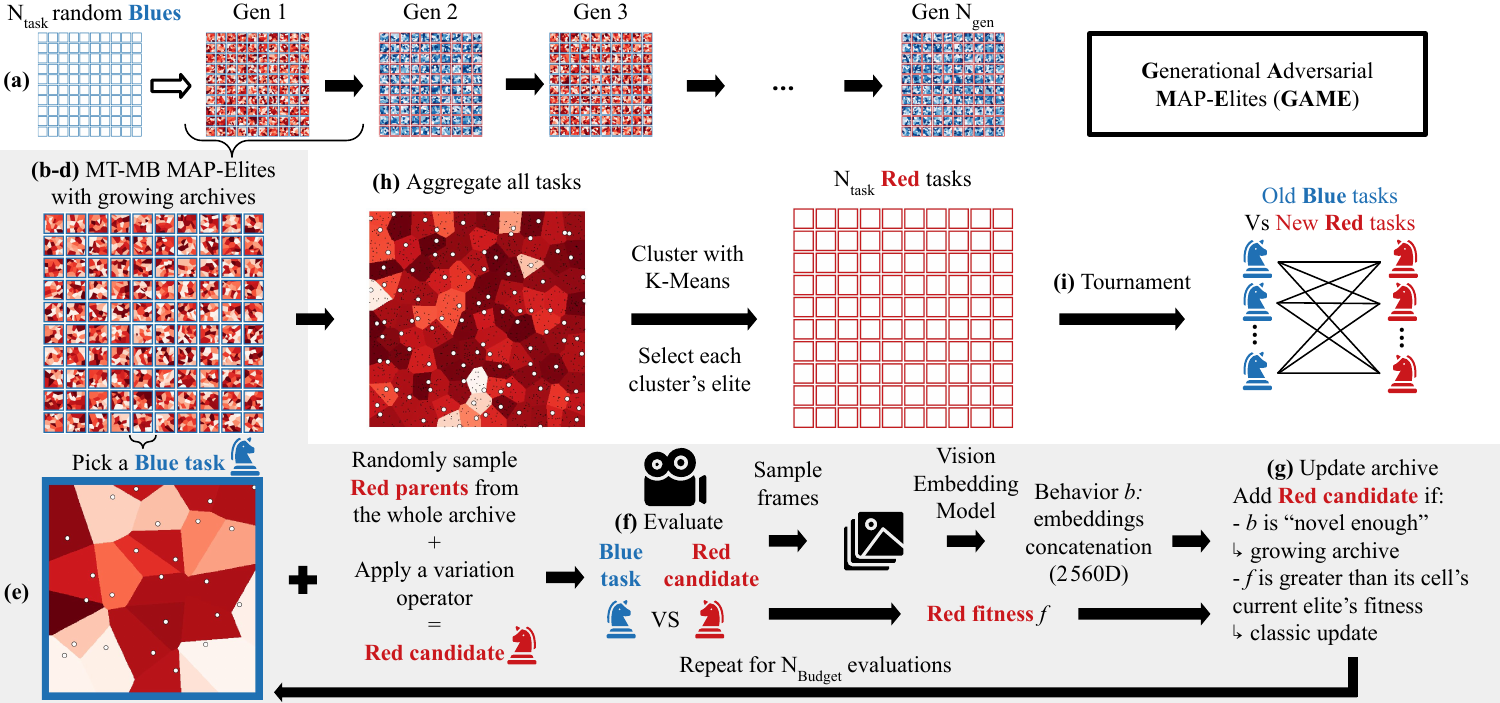}
    \vspace{-7.5mm}
    \caption{\textbf{GAME} is an adversarial coevolutionary QD algorithm that alternates the illumination of an adversarial problem using MTMB-ME~\cite{anne2023multi}, switching sides at each generation to promote arms race dynamics. A key feature is the ability to use a VEM as a domain-agnostic behavior space.}
    \label{fig:main_method}
   \vspace{-5mm}
\end{figure*}

One limitation of MAP-Elites is the need to define a behavior space. Automated Search for Artificial Life (ASAL)~\cite{kumar2024automating} has shown that a VEM, such as CLIP~\cite{radford2021learning}, can be used to explore the space of artificial life substrates. Inspired by this work, we propose, to our knowledge, the first use of a VEM as a behavior space for a MAP-Elites algorithm.

One difficulty with such a space is its high dimensionality (e.g., \num{2560} dimensions in this paper). The original MAP-Elites uses a grid to divide the behavior space, which is not scalable to high dimensions (as the number of cells grows exponentially with the dimensions). One solution is to use Centroidal Voronoi Tessellations (CVT)~\cite{vassiliades2016scaling} to partition the behavior space uniformly. However, in very high-dimensional spaces, it is difficult to predict the fraction of the space that the studied system covers, making it difficult to tune the selection pressure. If too many cells are available to fill, the pressure will be too low, reducing quality, and if there are not enough cells, the pressure will be too high, reducing diversity. One solution is to learn a low-dimensional representation using an autoencoder~\cite{cully2019autonomous}, which requires additional computation.

Our intuition is that we can grow the archive to cover the relevant region of the behavior space rather than defining it a priori. This is similar to how NSLC defines novelty using a growing archive of solutions \new{and is referred to as an unstructured archive. We unknowingly re-implemented a variant of Cluster-Elites~\cite{vassiliades2017comparison}}. Unlike NSLC, which uses a distance-threshold parameter to determine whether a new solution should be added to the archive, \new{Cluster-Elites} uses a predefined maximum number of cells and updates only their positions and boundaries. To do so, it updates the CVT centroids by replacing an old centroid with a new one if it reduces the minimum distance among all centroids, i.e., growing the archive. \new{An alternative method to handle a complex behavior space would be to use the Dominated Novelty Search~\cite{bahlous2025dominated}.
}

\section{Generational Adversarial MAP-Elites}

GAME is a QD algorithm that aims to illuminate both sides (Blue and Red) of an adversarial problem by leveraging coevolution (Fig.~\ref{fig:GAME_core_idea} and Fig.~\ref{fig:main_method}). \new{The main contribution of GAME is the coevolution of both sides of an adversarial problem through a sequence of generations by exploiting MTMB-ME~\cite{anne2023multi}'s ability to perform multi-task QD. The second contribution is the use of a VEM (CLIP~\cite{radford2021learning}) as a domain-agnostic behavior space in an algorithm of the MAP-Elites family, inspired by ASAL~\cite{kumar2024automating}, which leverages an unstructured archive~\cite{vassiliades2017comparison} to address high dimensionality.}

\subsection{Definitions}

\new{\textbf{Archive}: the set of solutions stored during evolution, organized in different cells which discretize the behavior space. A multi-task archive is a set of archives that use different fitness functions but share the same behavior and solution spaces.}

\new{\textbf{Behavior}: what allows us to determine whether two solutions do different things, e.g., armies moving with different patterns or robots with different morphologies.}

\new{\textbf{Cell}: a solution competes only with other solutions with similar behaviors; a cell defines a closed space in the behavior space that allows this local competition.}

\new{\textbf{Centroid}: a point in the behavior space that, when paired with a distance function and other centroids, allows for splitting the behavior space into cells, i.e., a solution is assigned to the cell whose centroid is the closest to its behavior.}

\new{\textbf{Elite}: the solution with the current highest fitness of its cell.}

\new{\textbf{Fitness}: the function to maximize. In an adversarial setting, this function depends on the opposing solution, i.e., the task.}

\new{\textbf{Solution}: a point in the search space. Both elites and tasks are solutions. A candidate solution can become an elite and be added to the archive if it is ``new enough'' or has higher fitness than the elite of its cell.}

\new{\textbf{Task}: defines a fitness and behavior function. In an adversarial setting, a task corresponds to a fixed solution of the opposing side against which candidate solutions are evaluated.}

\new{\textbf{Tournament}: the outcome of all possible pairs of evaluations between two opposing sets of solutions.}

\new{\textbf{Variation operator}: a combination of mutation and crossover operators specific to a search space that allows exploring the search space given one or more solutions.
}

\subsection{Main algorithm}

GAME is divided into two levels: (1) intergenerational illumination of solutions, alternating sides at each generation, Alg.~\ref{alg:main_alg} steps \textbf{a} to \textbf{d} and \textbf{h} to \textbf{i} \new{(the main contribution of this paper)} and (2) intragenerational illumination at each generation, using one side as tasks and the other as elites, Alg.~\ref{alg:MTMB-ME} steps \textbf{e} to \textbf{g} \new{(execution of MTMB-ME~\cite{anne2023multi})}. 

It proceeds as follows:
\begin{itemize}[leftmargin=*,align=left,widest={(c)}]
    \item[\textbf{(a)}] randomly sample $N_{task}$ Blue solutions \new{as initial tasks};
    
\noindent \textbf{For $N_{gen}$ generations:}
    
    \begin{itemize}[leftmargin=*]
        \item[\textbf{(b)}] alternate side if it is not the first generation;
        \item[\textbf{(c)}] initialize a multi-task multi-behavior growing archive with $N_{cell}$ for each task;
        \item[\textbf{(d)}] \new{bootstrap the archive with the evaluations from the previous tournament (if there has been a tournament);}
        
        \textbf{For $N_{budget}$ evaluations:}
        \begin{itemize}[leftmargin=*]
            \item[\textbf{(e)}] pick a $task$ at random and generate a candidate solution $s$: \new{if there are fewer than $N_{init}$ elites, use a random variation operator, otherwise randomly pick two elites from the whole archive and apply a variation operator specific to the search space};
            \item[\textbf{(f)}] evaluate $s$ against the $task$ to get the behavior descriptor $b$ and fitness $f$;
            \item[\textbf{(g)}] update the $task$'s archive (Alg.~\ref{alg:growing_archive}): (1) \new{if the distance between the new behavior $b$ and the centroids is larger than the distance between the closest pair of centroids, the candidate solution is added to the archive as a centroid and an elite, and one of the two centroids from the pair is removed; (2) else if its fitness $f$ is greater than its cell's elite's fitness, the candidate solution becomes the elite of its cell;}
        \end{itemize}
        \item[\textbf{(h)}] \new{aggregate the current side's elites}, cluster them into $N_{task}$ clusters using K-Means, and select the elite of each cluster, \new{which will form the new set of tasks for the next generation after alternating sides at step \textbf{(b)}};
        \item[\textbf{(i)}] \new{run a tournament between the previous and new tasks.}
    \end{itemize}
\end{itemize}

\vspace{-2pt}
\begin{algorithm} 
    \small
    \caption{GAME\\
    \textbf{Inputs}: \textcolor{red}{Red} search space \textcolor{red}{$S_{Red}$}, \textcolor{blue}{Blue} search space \textcolor{blue}{$S_{Blue}$} \\
    \textbf{Parameters}: $N_{gen}$, $N_{task}$ 
    }\label{alg:main_alg}
\begin{algorithmic}[1]
\State{} $Tasks \leftarrow$ Sample $N_{task}$ random \textcolor{blue}{Blue} solutions \Comment{\textbf{(a)}}
\State{} $Generations = [Tasks]$  \Comment{\new{Initialize with the first generation}}
\State{} $\mathcal{B} = \emptyset$ \Comment{For storing bootstrapping evaluations}
\For{$i=1$ \texttt{to} $N_{gen}$}
    \State{} $S \leftarrow$ \textcolor{red}{$S_{Red}$} if $i$ is odd else \textcolor{blue}{$S_{Blue}$} \Comment{\new{\textbf{(b)} - Alternate side}}
    \State{} $\mathcal{A} \leftarrow \operatorname{MTMB-ME}(Tasks, S, \mathcal{B})$ \Comment{\textbf{(c-g) - Alg.~\ref{alg:MTMB-ME}}}
    \State{} $Behaviors$ $\leftarrow$ Aggregate the archive's elites \Comment{\textbf{(h)}}
    \State{} Clusters $\leftarrow \operatorname{K-means}(Behaviors, k=N_{task})$ 
    \State{} $Tasks \leftarrow \{\operatorname{Elite}(cluster)\}_{cluster \in Clusters}$ 
    \State{} $Generations[i] = Tasks$ \Comment{\new{Store new generation}}
    \State{} $\mathcal{B} \leftarrow$ $Tasks$ vs $Generations[i-1]$ \Comment{\textbf{(i)} \new{- Tournament}}
\EndFor{}
\State{} \Return{} $Generations$
\end{algorithmic}
\end{algorithm}
\vspace{-10pt}

\begin{algorithm} 
    \small
    \caption{MTMB-ME with growing archive and a VEM\\ 
    \textbf{Inputs}: $Tasks$, search space $S$, bootstrap set $\mathcal{B}$\\
    \textbf{Parameters}: Number of evaluations $N_{budget}$, Number of initial random search $N_{init}$, evaluation function $\operatorname{Evaluate}$, variation operator $\operatorname{Variation}$, VEM 
    }\label{alg:MTMB-ME}
\begin{algorithmic}[1]
\State{} $\mathcal{A} \leftarrow$ Initialize $N_{task}$ growing archives \Comment{\textbf{(c)}}
\For{$(task, s, f, b)$ \texttt{in} $\mathcal{B}$} \Comment{\textbf{(d)} - Bootstrapping}
   \State{}  $\new{\mathcal{A}[task]} \leftarrow \operatorname{Update}(\mathcal{A}[task], s, f, b)$ \Comment{\textbf{Alg.~\ref{alg:growing_archive}}}
\EndFor{}
\For{$i=1$ \texttt{to} $N_{budget}$} \Comment{Main loop}
    \State{} $task \leftarrow$  Select a task at random from $Tasks$ \Comment{\textbf{(e)}}
    \If{ $\mathcal{A}$ has fewer than $N_{init}$ elites} \Comment{$N_{init}=100$}
        \State{} $s \leftarrow$ Sample a random solution from $S$ 
    \Else{}
        \State{} $s \leftarrow \operatorname{Variation\_Operator}(\mathcal{A})$ \Comment{\new{Specific to $S$}}
    \EndIf{}
    \State{} \new{$f, b \leftarrow \operatorname{Evaluate}(s, task)$} \Comment{\textbf{(f)}}
       \State{}  $\mathcal{A} \leftarrow \operatorname{Update}(\mathcal{A}[task], s, f, b)$ \Comment{\textbf{(g) - Alg.~\ref{alg:growing_archive}}}
\EndFor{}
\State{} \Return{} Archives
\end{algorithmic}
\end{algorithm}
\vspace{-10pt}

\begin{algorithm} 
    \small
    \caption{Growing unstructured archive update\\ 
    \textbf{Inputs}: archive $A$, solution $s$, fitness $f$, behavior $b$\\
    \textbf{Parameters}: max archive size $N_{cell}$, distance function $\operatorname{dist}$
    }\label{alg:growing_archive}
\begin{algorithmic}[1] 
\State{} $(C, E, E_{backup}) = A$  \Comment{Centroids, Elites, and Backup Elites}
\If{$\operatorname{size}(C) < N_{cell}$} \Comment{Add a new cell}
    \State{} $i = \operatorname{size}(E)$
    \State{} $C_i = b$ 
    \State{} $E[i] \leftarrow (s, f, b)$
    \State{} $E_{backup}[i] \leftarrow (s, f, b)$  
\Else{} \Comment{Check behavior and fitness}
    \State{} $distances = \{\operatorname{dist}(C_i, C_j)\}_{0 \le i < j < N_{cell}}$
    \State{} $d_{min} = \min(distances)$
    \State{} $d = \min\{\operatorname{dist}(b, C_i)\}_{0 \le i < N_{cell}}$
    \State{} $c_{id} = \operatorname{find\_cell}(C, b)$ \Comment{Closest centroid's index}
    \If{ $d > d_{min}$} \Comment{New enough behavior = growth}
        \State{} $j, k \leftarrow \operatorname{argmin}(distances)$  \Comment{\new{Find centroid to remove}}
        \State{} $d_j \leftarrow \min\{\operatorname{dist}(C_j, C_i)\}_{0 \le i \ne j < N_{cell} }$
        \State{} $d_k \leftarrow \min\{\operatorname{dist}(C_k, C_i)\}_{0 \le i \ne k < N_{cell} }$
        \State{} $k \leftarrow j $ if $d_j < d_k$ else $k$ \Comment{\new{Remove the closest to others}}
        \State{} $C_k \leftarrow b$
        \State{} $E[k] \leftarrow (s, f, b)$
        \State{} $E_{backup}[k] \leftarrow (s, f, b)$
        \For{$i=0$ to $N_{cell}$} \Comment{Check and repair holes}
            \If{$\operatorname{find\_cell}(C, E[i].b) \ne i$}
                \State{} $E[i] \leftarrow E_{backup}[i]$
            \EndIf{}
        \EndFor{}
    \ElsIf{f $>$ E[$c_{id}$].f} \Comment{Better fitness}
        \State{} $E[c_{id}] \leftarrow (s, f, b)$
    \EndIf{}
\EndIf{}
\State{} \Return{} $(C, E, E_{backup})$
\end{algorithmic}
\end{algorithm}

\subsection{Implementation details}



\subsubsection{Using a VEM for behavior space}

\new{During step \textbf{(f)}, GAME: (1) collects the video of the evaluation, (2) subsamples it to get $(frame_i)_{1:i:f}$, (3) passes each frame through the VEM (CLIP) to obtain the embeddings $(E_i)_{1:i:f}$ where each $E_i\in[-1, 1]^{256}$, and concatenates them in one vector $b=\operatorname{concatenation((E_i)_{1:i:f})}\in{[-1,1]^{f\times256}}$. GAME follows ASAL by using cosine similarity to compute the distance between two such behaviors (which are \num{2560}D in this paper). The reason is that the Euclidean distance becomes less informative as the dimensionality increases. This does not affect the growing archive, as it uses relative comparisons of distances between behaviors rather than their actual positions.}

\subsubsection{Growing an unstructured archive}

\new{When replacing a centroid with a new behavior, GAME needs to remove an existing centroid. Currently, it selects the one from the closest pair of centroids that is closest to the other centroids. Removing one of the two from the pair ensures that the minimal distance between two centroids is strictly increasing. Preliminary experiments on a toy problem showed that removing the one closest to the others slightly increased the resulting Coverage and QD-Score. In all our experiments, the computational bottleneck was the evaluation, so the additional computational cost of comparing the distances between the two centroids was negligible.}

When a new behavior takes over a cell, it can ``steal'' the elites of neighboring cells, thereby altering the entire CVT and creating ``holes''. A simple solution to fill those ``holes'' is to reinstate the centroid as the elite, since it cannot be ``stolen.'' This requires storing an additional solution per cell. More elaborate heuristics can be employed, e.g., storing all past elites and checking against the current CVT. However, we found that this does not significantly improve performance while increasing computational and memory requirements.

\new{In terms of complexity, deciding if the new behavior is to be added requires computing all the pairwise distances, $O(N_{cell}^2)$, and repairing the holes requires looping through all cells to determine whether its centroid is reassigned to the new cell, which is also $O(N_{cell}^2)$. So the overall complexity of updating the archive is $O(N_{cell}^2)$. In practice, we used vectorized NumPy functions, which reduce runtime.}

\new{To implement the growing archive, we unknowingly re-implemented a variant of the unstructured archive Cluster-Elites~\cite{vassiliades2017comparison} with the differences that (1) it uses a fixed archive size (while Cluster-Elites uses a schedule to gradually increase the number of cells); (2) it updates the archive one new evaluation at a time (while the Cluster-Elites updates it after a batch of evaluations); and (3) it prunes the centroid with the closest distance to the others (while the Cluster-Elites uses a mean distance to the $d+2$ nearest neighbors, where $d$ is the dimensionality of the behavior space).
}

\subsubsection{Sharing solutions between generations}

\new{GAME handles the interdependence of the behavior and fitness between the two opposing solutions by using multi-task archives, i.e., each elite competes only against elites evaluated against the same opposing solution. To add a solution to the archives, it needs to be evaluated against at least one current task. Because the tasks change at each generation, to share information between two successive generations on the same side, $i-2$ and $i$, GAME must re-evaluate the elites of generation ${i-2}$ against the tasks of generation ${i}$. Fully sharing all elites would require a significant number of evaluations (i.e., $N_{cell}\cdot N_{task}^2$). To balance the evaluations between sharing information from the previous generation and searching for new solutions, we use a bootstrap tournament that only evaluates the tasks of generation $i-1$ (which are a subset of elites from generation ${i-2}$ that preserve diversity and quality) against the new tasks of generation ${i}$ (i.e., $N_{task}^2$ evaluations).}


\section{Case Study: Parabellum}

\begin{figure*}[th]
    \centering
    \includegraphics[width=\linewidth, trim=0cm 6.5cm 0cm 0cm]{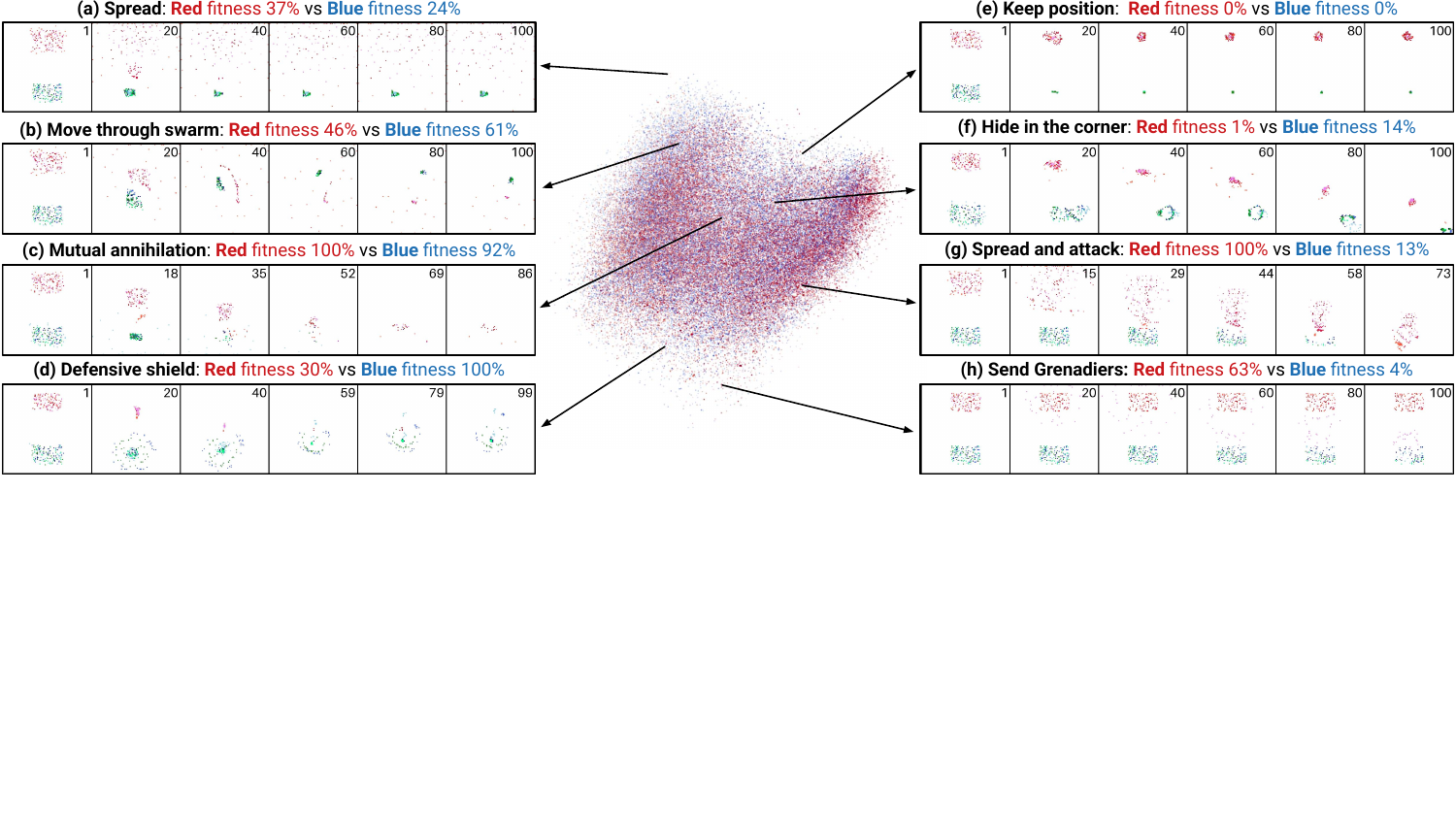}
    \caption{\textbf{GAME's illumination of a multi-agent adversarial game.} The point cloud is a 2D PCA projection (\num{22}\% and \num{10}\% explained variance) of the intergenerational tournament between elites found for one run of GAME across \num{20} generations with $\num{100000}$ evaluations per generation. We display timed snapshots of eight duels exhibiting different behaviors. We also indicate the fitness of both sides, represented as the percentage of the opposing side's depleted health. Videos of these duels and supplementary data are available in the repository \url{https://github.com/Timothee-ANNE/GAME} (video figure\_3.mp4). }
    
    \label{fig:teaser}
    \vspace{-4mm}
\end{figure*}

We first evaluate GAME \new{and its ablations} in the battle game research environment Parabellum~\cite{anne2025harnessing}. Battle games are inherently adversarial, making them a direct target for illuminating adversarial problems. In addition, they employ many agents in a top-down view, making behavioral comparison non-trivial. Parabellum also shares similarities with ALife simulations, which inspired us to apply the same VEM as in ASAL~\cite{kumar2024automating}. 

\subsection{Parabellum Multi-Agent Game}

Parabellum is a multi-agent battle game in which two sides, Blue and Red, aim to eliminate the other, i.e., the fitness of each side is the sum of the depleted health of the opposing side's units. Each unit can only see other units within its sight range (\num{15}~m, with the map being \num{100}~m wide). At each time step, each unit can either move, stand, attack an enemy in reach, or heal an ally in reach, given its local observation (position, type, side, and health of all the units in sight). It is implemented in JAX~\cite{jax2018github}, a Python library for just-in-time compilation and vectorization. It permits fast parallelization of the units' behaviors using a dedicated GPU. Parabellum contains stochasticity, but we use the same seed for all encounters to make comparisons fair.

We define five types of units (each with a different color for visualization):
\textbf{spearman}: slow and high-health close combat unit;
\textbf{archer}: low-health long-range combat unit;
\textbf{cavalry}: fast and medium-health close combat unit;
\textbf{healer}: low-health healing unit with close range;
\textbf{grenadier}: low-health mid-range unit that inflicts area damage to all units.

Each side comprises \num{32} units of each type uniformly spread in each side's starting area, which are placed so that no unit initially sees an enemy unit (see Fig.~\ref{fig:teaser} frames 1).

\subsection{Behavior Tree as Controller}
Parabellum uses behavior trees (BTs)~\cite{colledanchise2018behavior} to control each unit. BTs are commonly used in robotics and video games, intuitive to design, and inherently interpretable. 

More specifically, the units on each side share the same BT, so the search space of GAME is one BT for the Red side and one BT for the Blue side. At each time step, each unit visits its BT starting from the root node and performs a leftmost depth-first traversal of the tree until it finds a valid action. The BT comprises intermediate nodes: Sequence/Failwith nodes, which stop at the first invalid/valid node, and leaf nodes: Action nodes that return an action and its validity, and Condition nodes that return a boolean. To enable fast evaluations, we implemented a BT evaluation function in JAX that allows vectorization of the BT evaluation, albeit at the cost of setting a maximum number of leaves (100 in this paper). 

Their tree structure allows the use of genetic variation operators. For example, \cite{iovino2021learning} evolve BTs for robot control, and \cite{montague2023quality} use MAP-Elites with BTs to learn controllers for robot swarms. 
Taking inspiration from them, \new{Parabellum's \textbf{variation operator} randomly selects one of the following mutations}: deleting a sub-tree (\num{35}\%), adding a random node at a random location (\num{21}\%), mutating a node by changing its parameters (\num{7}\%), replacing a node with a random node (\num{7}\%); or a crossover, i.e., copying a random sub-tree of another BT at a random location (\num{30}\%). Those probabilities could be tuned, but they do not appear to significantly affect GAME's performance. Another advantage of using BTs is that it is straightforward to increase their complexity by allowing them to grow, promoting open-ended evolution~\cite{harrington2019escalation}. Such an increase in complexity is challenging to achieve with neural networks and requires specialized methods, e.g., NEAT~\cite{stanley2002evolving}. 

The BT evolution requires designing atomic functions (Actions and Conditions) that take the unit's local observation and return the corresponding output. To allow synchronization between units, each side can set a target position on the map and move toward it even if it is out of sight. Different qualifiers parametrize the atomic functions to improve interpretation and usage, e.g., a \textit{target} qualifier that can be set to closest, farthest, weakest, strongest, or random. 

The available Actions are: 
\texttt{Stand} (do nothing); 
\texttt{Attack} (attack the \textit{target} enemy in reach of a given type); 
\texttt{Heal} (heal the \textit{target} ally in reach of a given type);
\texttt{Move} (move toward/away from the \textit{target} ally/enemy of a given type);
\texttt{Go To} (go to the target position); 
\texttt{Set Target} (mark the position of the \textit{target} ally/enemy as the target position).

The available Conditions are:
\texttt{In Sight} (Is there an ally/enemy of a given type in sight?)
\texttt{In Reach} (Is there an ally/enemy of a given type in reach?)
\texttt{Is Dying} (Is my health or an ally's/enemy's health below a given threshold?)
\texttt{Is Type} (Am I of a given type?)
\texttt{Is Set Target} (Is the target position set on the map?)

\subsection{GAME variants and ablations}

To evaluate GAME, we evaluate seven variants:
\begin{itemize}
    \item \textbf{GAME-SO}: full variant that uses a single-objective fitness to maximize the opposing side's depleted health;
    \item \textbf{GAME-MO}: full variant with a multi-objective fitness that first maximizes the opposing side's depleted health and secondarily minimizes the size of the BT;
    \item \textbf{GAME-SO (no bootstrap)}: GAME-SO with an ablation of the bootstrapping between generations;
    \item \new{\textbf{GAME-SO (Positions)}; GAME-SO with the concatenations of the units' positions at 4 timesteps (25, 50, 75, and 100) instead of a VEM (it has the same number of dimensions, i.e., $\num{2560}=4\times 320 \times 2$);}
    \item \textbf{GAME-SO (Handcrafted)}: GAME-SO with the average remaining health and the time before completion (i.e., one side wins or \num{100}~steps timeout) as a 2D handcrafted behavior space instead of a VEM;
    \item \textbf{Diversity-only}: do not use fitness, only rule (g.1);
    \item \textbf{Quality-only}: MT-ME instead of MTMB-ME, i.e., each task has only one solution.
\end{itemize}

We run three replications of $N_{gen} = \num{20}$ generations for each variant with $N_{budget}=\num{100000}$ evaluations of $\num{100}$ steps per generation, $N_{task}=\num{100}$ tasks per generation, and $N_{cell}=\num{25}$ cells per task's archive.

\subsection{Measures}

\begin{figure*}[ht]
    \centering
    \includegraphics[width=\linewidth]{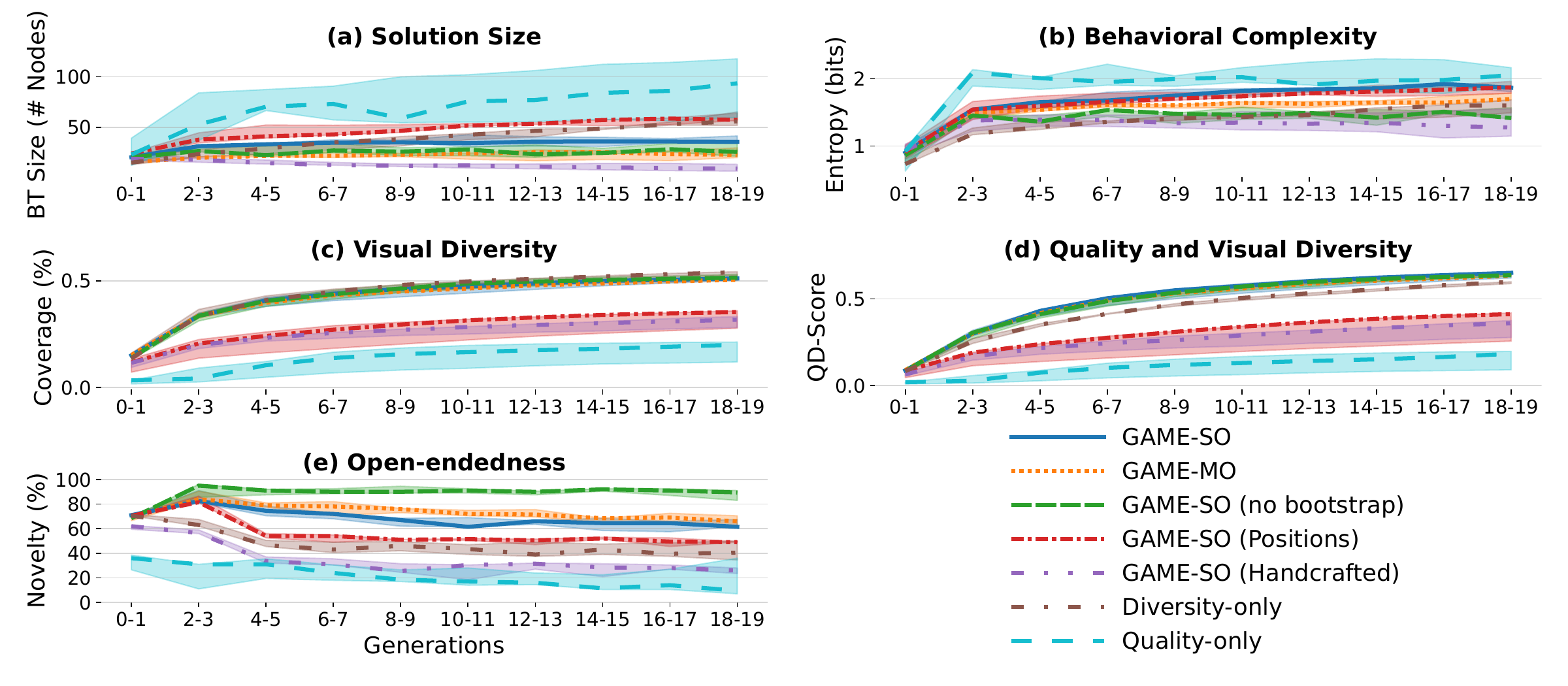}
    \vspace{-10mm}
    \caption{\textbf{Variants and ablations comparisons.} The solid line is the median, and the shaded area the min and max of \num{3} replications over \num{20} generations. \new{(a-b) Larger BTs do not necessarily lead to higher complexity, as GAME-SO has the 2nd-highest behavior complexity but the 4th-highest BT size. (c–d) GAME requires the VEM to reach the highest diversity and QD-Score. (e) Removing bootstrapping leads to a constant discovery of novel solutions.} }
    \vspace{-3pt}
    \label{fig:5comparisons}
\end{figure*}

After each run, we conduct a tournament between the tasks of each generation, \num{10} Blue generations against \num{10} Red generations (i.e., \num{1000} Red elites against \num{1000} Blue elites), yielding \num{1000000} evaluations. This allows us to study whether there are intergenerational improvements.

\paragraph{Solution Size and Behavioral Complexity}
To measure variations in solution size, we compute the number of nodes of the BT elites (Fig.~\ref{fig:5comparisons}.a), and for behavioral complexity, we measure the entropy of the distribution of actions performed over time and units of the same side (Fig.~\ref{fig:5comparisons}.b).

\paragraph{Visual Diversity and Quality Visual Diversity}

\begin{figure}[ht]
    \centering
    \includegraphics[width=\linewidth]{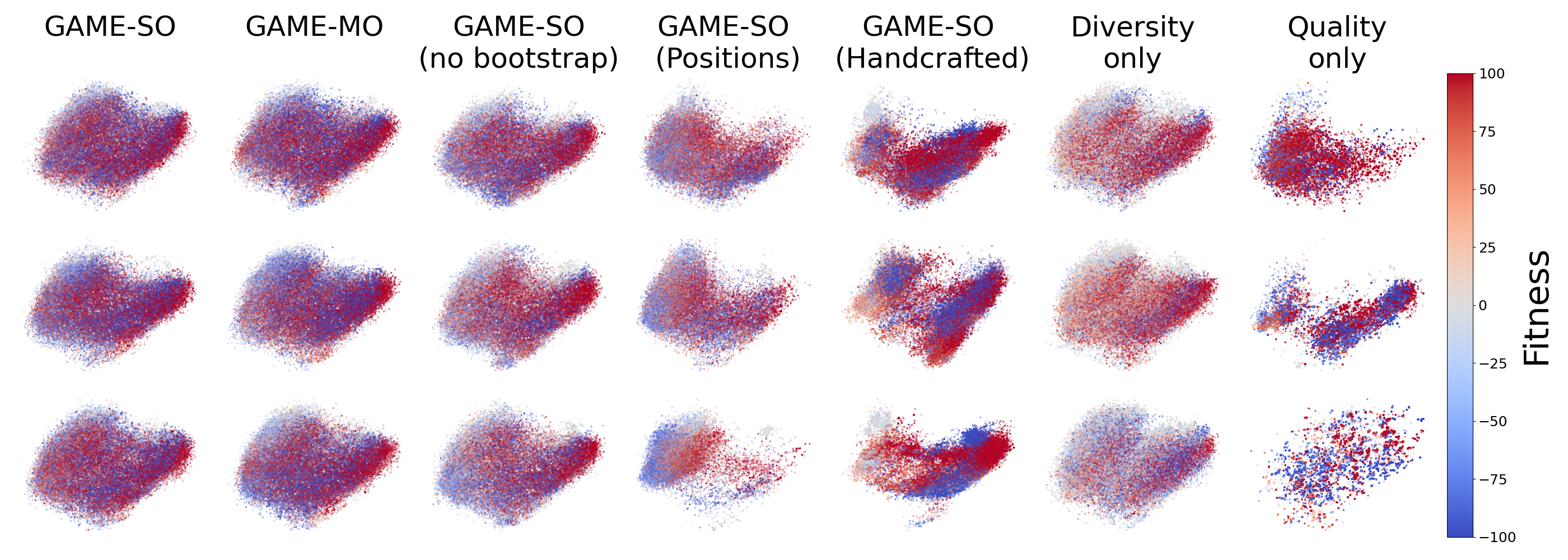}
    \vspace{-8mm}
    \caption{\textbf{PCA projections of the intergenerational tournaments' behaviors for each replication of each variant.} GAME variants with a VEM show the most homogeneous coverage with less variance between replications.}  
    \label{fig:pcas} 
    \vspace{-0.1cm}
\end{figure}

Coverage (the proportion of non-empty cells, Fig.~\ref{fig:5comparisons}.c) and QD-Score (the average fitness of the elites, Fig.~\ref{fig:5comparisons}.d) are classic measures of QD algorithms, but are not easily computed in a large behavior space. Following \cite{dharna2024quality}, we use a single 2D PCA projection for all the behaviors of the intergenerational tournaments for all replications (Fig.~\ref{fig:pcas}), discretize this space with a $100\times100$ grid, and compute the corresponding elites. The two components capture only \num{22}\% and \num{10}\% of the explained variance, which limits the measure's validity but still allows discrimination between variants. An autoencoder could capture more variance at the price of additional computation~\cite{cully2019autonomous}.

\begin{figure}[ht]
    \centering
    \includegraphics[width=\linewidth]{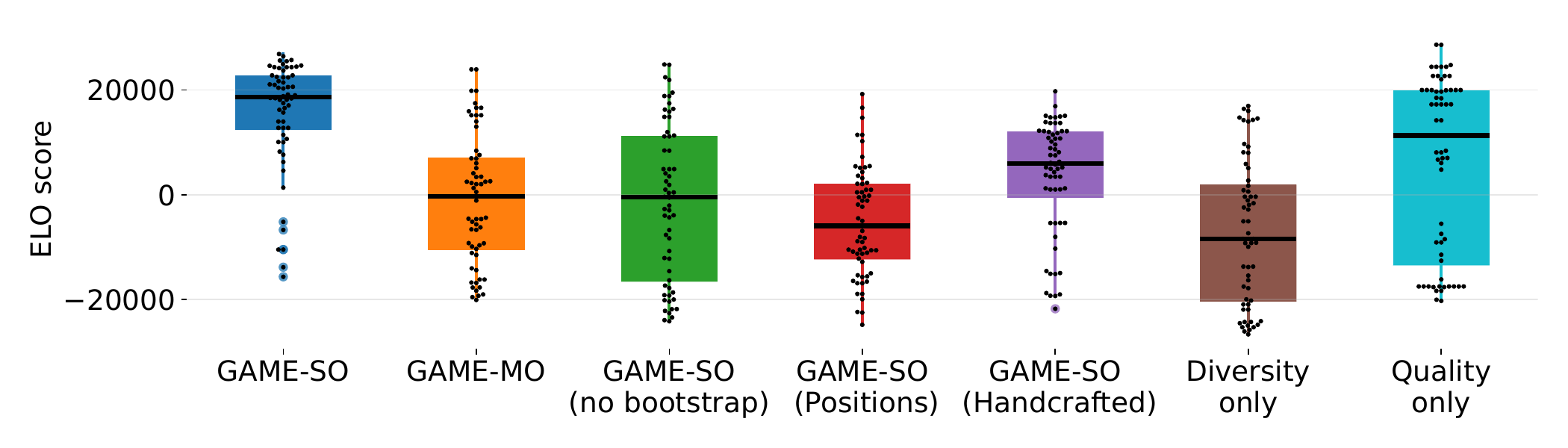}
    \vspace{-8mm}
    \caption{\textbf{Tournament ELO score between each replication's \num{10} best solutions.} \textbf{GAME-SO} is significantly better than all variants but \textbf{Quality-only} (p-value $ <0.001$ with a Mann-Whitney U test and Bonferroni correction). }
    \label{fig:elo_score}
    \vspace{-3pt}
\end{figure}

\paragraph{Quality}
One limitation of computing quality directly from fitness is that it can be high because the opposing side is weak, not because the winning side is strong. To measure a less ambiguous quality of the elites found by each method, we follow \cite{dharna2024quality} and select the best \num{10} solutions from each side for each run of each variant, perform a round-robin tournament, and compute the ELO score~\cite{elo1978rating} (Fig.~\ref{fig:elo_score}). 

\paragraph{Open-endedness}
Coverage measures the volume of novelty discovered by GAME with the VEM. One limitation is that this behavior is dependent on two solutions. Another limitation is that visual behavior may not capture the full range of novelty. We use another measure similar to PATA-EC proposed in Enhanced-POET. The idea is that a BT is different from another if, in a tournament, the ranking it generates in the opposition is different. Using the intergenerational tournament, we compute the rankings of all BTs and count the number of new rankings in each generation relative to the previous generation (Fig.~\ref{fig:5comparisons}.e). The idea is that an open-ended system will continually create challenges that require new solutions.   

\subsection{Results}

\paragraph{Solution Size (Fig.~\ref{fig:5comparisons}.a)} 

\textbf{Quality-only} has a significantly higher increase in BT size. This can be explained by the lack of diversity, which narrows the optimization to a smaller set of solutions (one per task, rather than $N_{cell}=25$), thereby allowing more mutations to accumulate. \new{\textbf{Diversity-only} and \textbf{GAME-SO (Positions)} have a consistently increasing BT size. For \textbf{Diversity-only}, it can be explained by the absence of fitness regulation, leading to preserving ``bad'' nodes that would have been removed otherwise. For \textbf{GAME-SO (Positions)}, it can be explained by a behavior space that is harder to expand, thus requiring the accumulation of more mutations.} 

\paragraph{Behavioral Complexity (Fig.~\ref{fig:5comparisons}.b)} 

\textbf{Quality-only} also has the highest increase in behavioral complexity, with the same explanation as before. What is more interesting is the presence of \textbf{GAME-SO} as a close second. Compared to \textbf{GAME-MO}, this suggests that minimizing BT size prunes necessary stepping stones that lead to more complex behaviors. \textbf{Diversity-only} is ranked fifth, showing that increased BT size does not necessarily lead to greater behavioral complexity.

\paragraph{Visual Diversity (Fig.~\ref{fig:5comparisons}.c and Fig.~\ref{fig:pcas})} 

\textbf{Quality-only}, \textbf{GAME-SO (Handcrafted)}, \new{and \textbf{GAME-SO (Positions)}} exhibit significantly less diversity than other variants, which show similar diversity levels, with \textbf{Diversity-only} being slightly superior. \new{\textbf{GAME-SO (Positions)}, \textbf{GAME-SO (Handcrafted)}, and \textbf{Quality-only} also show more variance between runs in the PCAs projections, and sparser and more heterogeneous covering of the behavior space. Those results validate the use of a VEM as a behavior space.} 

\paragraph{Quality and Visual Diversity (Fig.~\ref{fig:5comparisons}.d)} 

\textbf{Quality-only}, \textbf{GAME-SO (Handcrafted),} \new{and \textbf{GAME-SO (Positions)}} show significantly worse QD-Scores (due to their poorer coverage). \textbf{Diversity-only} shows only a slightly worse QD-Score, suggesting that diversity alone is already a powerful optimizer for GAME, though the best QD-Score is achieved when also considering quality.

\paragraph{Quality (Fig.~\ref{fig:elo_score})} 

Comparing the quality of the best solutions, we observe that \textbf{GAME-SO} and some instances of \textbf{Quality-only} produce significantly better BTs (p-value $ <0.001$ with a Mann-Whitney U test and Bonferroni correction). This suggests that increased behavioral complexity yields significant performance improvements, and that directly attempting to reduce solution size as a secondary objective may prevent the discovery of the highest-performing solutions. \textbf{Quality-only} has the largest variance between replications. This result suggests that diversity, quality, the VEM, and the bootstrap phase are all necessary to discover the best BTs.

\paragraph{Open-endedness (Fig.~\ref{fig:5comparisons}.e)} 

\textbf{GAME-SO} and \textbf{GAME-MO} show similar open-endedness, which slightly decreases with each iteration, finding \num{80}\% new tasks in the 2nd and 3rd generations and declining to \num{60}\% by the final generations. \textbf{GAME-SO (no bootstrap)} consistently generates \num{90}\% novel BTs, suggesting this variant is the only one showing signs of true open-endedness throughout all \num{20} generations, but also that bootstrapping may be detrimental to open-endedness. The other variants exhibit less open-endedness, with \textbf{Quality-only} approaching \num{0}\% novelty. This further demonstrates that diversity prevents getting trapped in local minima.

\paragraph{\new{Robustness to fixed random seed}}

\new{To assess whether GAME overfitted to the random seed of this stochastic environment, we recomputed the intergenerational tournament for the first replication of \textbf{GAME-SO} with a different seed. The average fitness change is $0.04$, and the average absolute fitness change is $0.18$, indicating that fitness fluctuates; however, because it affects both sides, the average fitness does not change significantly. The Spearman rank-order correlation coefficient between the two resulting ELO score rankings is 0.963 (p-value $< 0.001$) for the 998 Red elites and 0.961 (p-value $< 0.001$) for the 999 Blue elites. This high correlation shows that GAME is robust to the fixed random seed. 
}

\paragraph{Two illumination takeaways for Parabellum}

\begin{figure}
    \centering
    \includegraphics[width=0.9\linewidth]{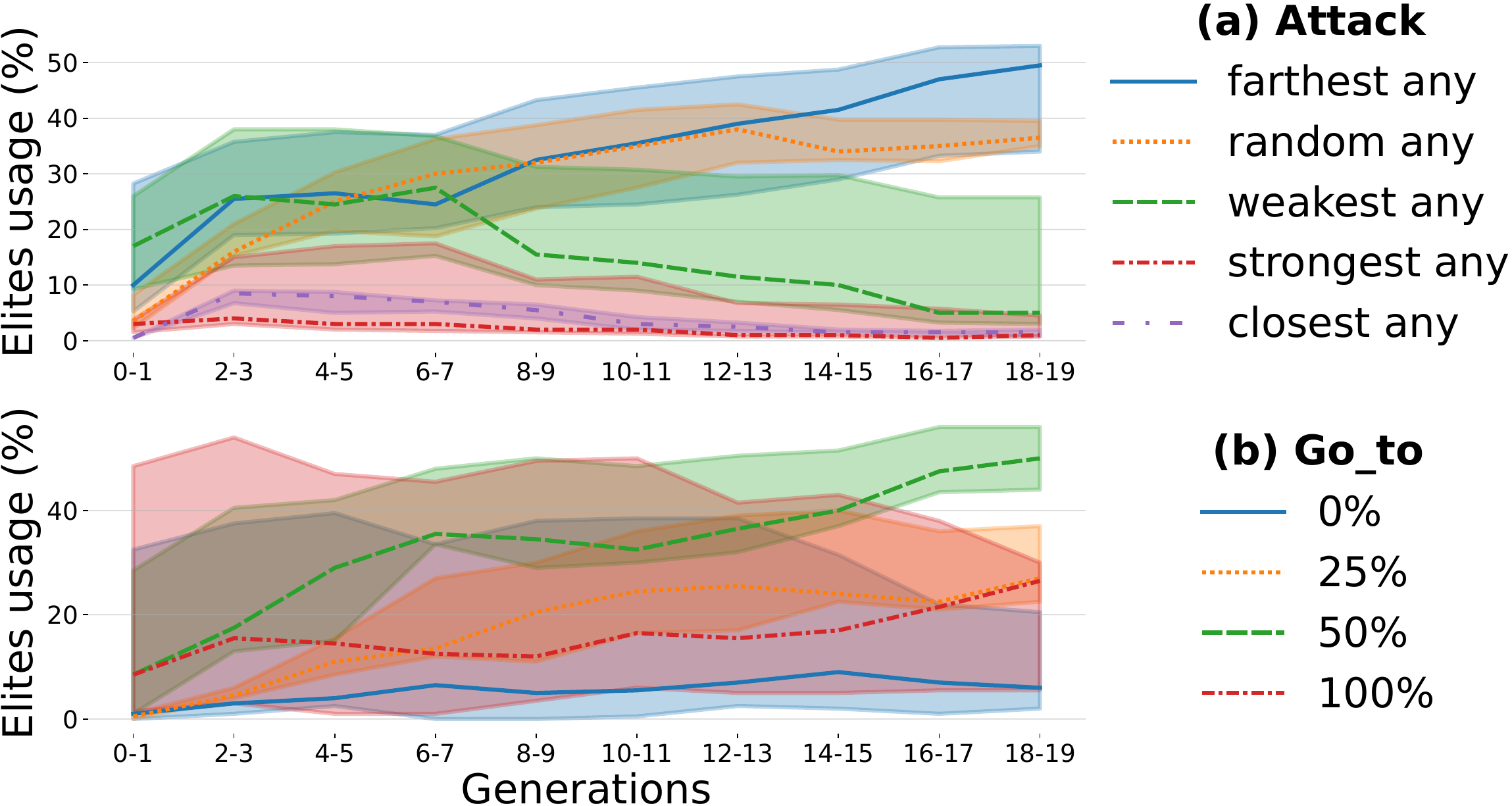}
    \vspace{-3mm}
    \caption{\new{\textbf{Parabellum global trends}: Percentage of elites using (a) the \texttt{Attack} atomic with different \textit{targets} and (b) the \texttt{Go\_to} atomic for different \textit{thresholds} through the generations. The solid line represents the median, and the shaded area between the min and max of \textbf{GAME-MO}'s three replications.}}
    \label{fig:global_trends}
    \vspace{-2mm}
\end{figure}

\new{We examined the atomic usage of \textbf{GAME-MO}'s three replications to identify global trends in Parabellum and found two takeaways.} 

\new{Regarding the \texttt{Attack} atomic, all replications converged to favor attacking either the farthest unit or a unit at random (Fig.~\ref{fig:global_trends}.a). Attacking at random spreads damage and avoids wasting resources on overkills. Attacking the farthest unit is also beneficial because grenadiers deal area damage that can be avoided if they target a unit sufficiently far away.}

\new{Examining the \texttt{Go\_to} atomics, all replications favor using a threshold distance equal to \num{50}\% of the sight range while avoiding exact positioning, i.e., \num{0}\% (Fig.~\ref{fig:global_trends}.b). This likely represents a balance between spending time moving precisely to the target position (i.e., \num{0}\%), which prevents units from performing other actions, and merely moving within sight of the target (i.e., \num{100}\%), which limits visibility of what can be observed from the target position itself.}

\paragraph{An example of arms race}

\begin{figure}[ht]
    \centering
    \includegraphics[width=\linewidth]{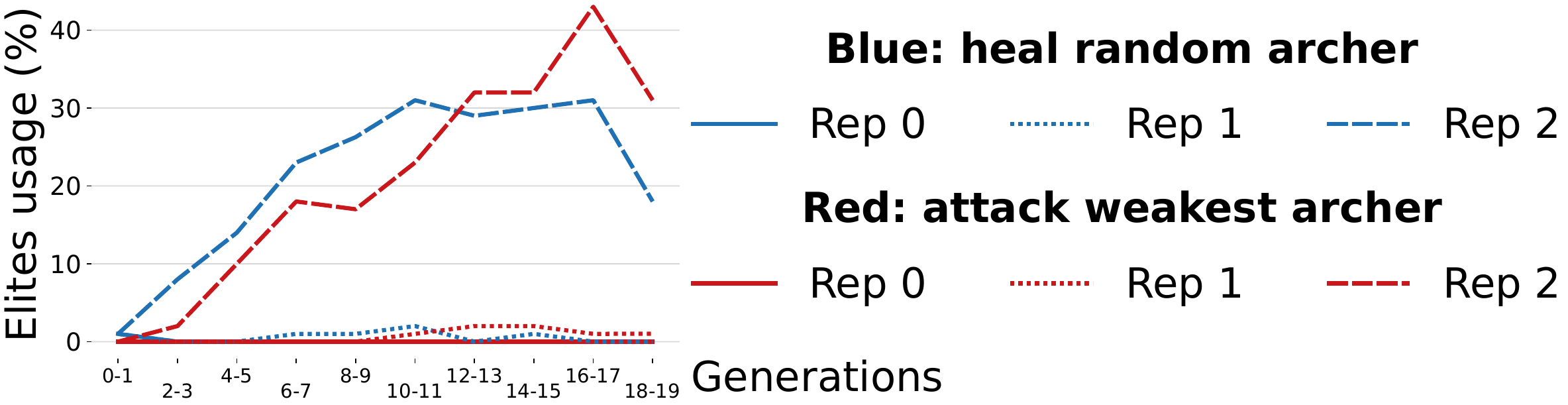}
    \vspace{-8mm}
    \caption{\textbf{Arms race example}: Percentage of Red elites using the \texttt{Attack} atomic on the weakest enemy archer and Blue elites using the \texttt{Heal} atomic on random ally archers through the generations for \textbf{GAME-SO}'s three replications. For replication \num{2}, as more Red elites targeted the Blue archers, Blue elites evolved to heal their archers. }
    \label{fig:arms_race_example}
    \vspace{-0pt}
\end{figure}

We found an arms race in one of \textbf{GAME-SO}'s replications (Fig.~\ref{fig:arms_race_example}). \new{Looking at the activation of different atomics of the evolved behavior trees, we found that among all Red elites, there was an increasing use of the \texttt{Attack weakest archer} atomic from 0\% at generation 0 to 40\% at generation 16 and at the same time, among Blue elites there was a similar increase in usage of the \texttt{Heal random archer} atomic from 1\% at generation 1 to 30\% at generation 17. These two atomics were not significantly used in the two other replications, which could indicate that the continuous increase of both atomics is a result of a particular arms race between the two sides, as archers are the most fragile but deadly units. } 

\subsection{Discussions and Future Work}
We hypothesized that minimizing BT size as a secondary objective would not compromise quality and would yield smaller, more interpretable BTs. However, the results indicate that it prunes neutral mutations that appear to be essential stepping stones for high-performing solutions. This supports the neutral theory of molecular evolution~\cite{kimura1979neutral}, which posits that most variation at the molecular level is neutral yet can drive the evolution of complex organisms in nature.

We also hypothesized that bootstrapping each generation using solutions from previous ones would accelerate the search compared to starting from scratch. The results confirm this hypothesis. Nonetheless, \textbf{GAME-SO (no bootstrap)} is the only variant demonstrating constant generation of new solutions throughout all \num{20} generations. This phenomenon is related to extinction events~\cite{lehman2015enhancing} and warrants further investigation, as it currently does not yield the best diversity or quality. 

One limitation of this case study is the use of BTs as controllers. They are inherently interpretable but require handcrafted atomics that limit the possible behaviors. An interesting future direction is to use end-to-end neural networks that map observations to actions, potentially employing neuroevolution~\cite{stanley2002evolving}, to move toward a more open-ended search space.

\section{Case study: Wrestling}

\begin{figure*}[ht]
    \centering
    \includegraphics[width=\linewidth]{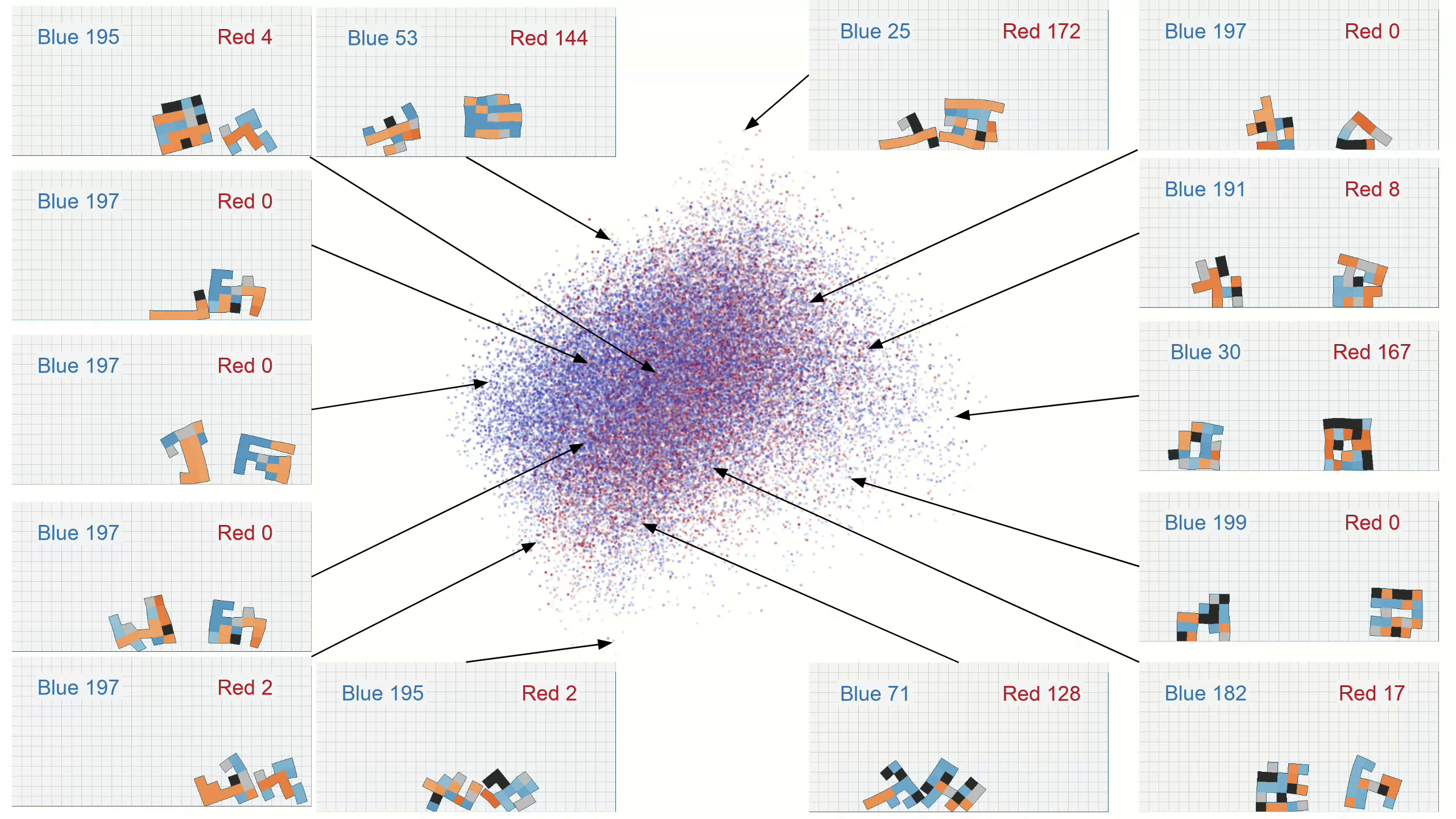}
    \vspace{-0.7cm}
    \caption{\textbf{Wrestling illumination with sampled snapshots}. The central point cloud is the 2D PCA of the visual embedding of one illumination from \textbf{GAME} of the Wrestling adversarial problem (capturing 13.4\% and 8.4\% of the variance). The outer edge snapshots show examples of frames captured at the end of a diverse set of evaluations. The PCA captures slow-moving robots on the right and faster-moving robots on the left, each with a large variety of morphologies. The numbers correspond to the final fitness of each solution (maximum is \num{200}).
    See Figure\_9.mp4 at \url{https://github.com/Timothee-ANNE/GAME}.
    }
    \label{fig:example_morphologies_pca}
    \vspace{-0.2cm}
\end{figure*}

\subsection{Environment}
\new{To show the generality of GAME with a VEM, we apply it to a different domain: 2D soft robots.} Replicating the biological evolution accomplishment of creating the diversity of living creatures' morphology has been a long-standing research topic in the artificial life community~\cite{langton1997artificial}.
An early work involved evolving articulated rigid-body creatures in an adversarial problem in which each creature's goal is to get as close as possible to a central cube~\cite{sims1994evolving}. To more closely resemble real creatures, recent work has focused on soft robots for locomotion~\cite{cheney2014unshackling, kriegman2017minimal, cheney2018scalable, bhatia2021evolution, mertan2023modular, nadizar2025enhancing}.

While the coevolution of morphology and its control is an interesting and challenging topic~\cite{cheney2018scalable, bhatia2021evolution, mertan2023modular, nadizar2025enhancing}, it is beyond the scope of this introductory GAME paper to explore the intersection of adversarial coevolution and body–brain coevolution. We instead focus on the adversarial coevolution of morphology in an adversarial environment, where the morphology passively provides actuation. We build a custom environment using EvoGym~\cite{bhatia2021evolution} for 2D soft-robot simulation, which we refer to as Wrestling. We follow \cite{cheney2014unshackling, kriegman2017minimal} by defining passive and active voxels that change surface following a sine wave pattern. We defined seven types of voxels represented by an integer: (0) empty (transparent), (1) rigid (black), (2) soft (gray), (3) horizontal actuated in-phase (orange), (4) vertical actuated in-phase (royalblue), (5) horizontal actuated anti-phase (gold), and (6) vertical actuated anti-phase (skyblue).

We set the sine-wave period to \num{12} timesteps because it empirically resulted in most randomly sampled robots exhibiting non-stationary behavior. The \textbf{solution space} is $5\times5$ 2D soft robots, \new{i.e. $\{0, 1, 2, 3, 4, 5, 6\}^{25}$,} with the constraints that there is at least one actuated voxel and that all voxels are connected. See Fig.~\ref{fig:example_morphologies_pca} for examples of different morphologies. 
The \textbf{variation operator} randomly chooses between adding, deleting, or mutating a voxel ($\frac{1}{3}$ for each) while respecting the constraints and does this $k$ times ($k=3$ in this paper).

In Wrestling, two robots start symmetrically at the edge of an arena 30 voxels wide. The \textbf{fitness function} is the percentage of timesteps the robot was closest to the center after \num{200} timesteps. This leads each robot to reach the center as quickly as possible, push the other, and resist the other's pushes. One specificity of this fitness function compared to the one used in Parabellum is that it is a relative measure of quality, i.e., given two competing blue and red solutions, $s_{blue}$ and $s_{red}$, $f(s_{blue}) = 1 - f(s_{red})$. This means that even trivially bad solutions (e.g., standing morphologies) would make at least one of the two opposing fitness positive. This prevents evaluating quality directly using the QD-Score and instead requires using a tournament or another quality measure (e.g., solution velocity). This is interesting because one-sided illumination would require knowing a high-quality solution to avoid trivially overfitting to a bad solution, \new{ which we illustrate by comparing GAME to a one-sided illumination baseline.} 

The \textbf{behavior space} is, similarly to Parabellum, the concatenation of the visual embedding of CLIP~\cite{radford2021learning} for five frames of the video. For speed optimization, we do not use EvoGym's visualization but instead use a JAX reimplementation that directly produces RGB arrays for CLIP.

\subsection{Variants and Ablations}

To demonstrate the generality of GAME, we apply the same implementation used on Parabellum to Wrestling but with a smaller evaluation budget ($N_{budget}=\num{20000}$, $N_{task}=50$, $N_{gen}=10$, and $N_{cell}=20$). We performed $\num{3}$ replications for each of these 4 variants and baselines: 
\begin{itemize}
    \item \textbf{GAME}: full variant;
    \item \new{\textbf{GAME-CVT}: ablation that uses a fixed CVT archive of size $N_{cell}=20$ for each task created at initialization;}
    \item \new{\textbf{MTMB-ME}: one-sided illumination that uses MTMB-ME with the same total number of evaluations but during only one generation against a fixed set of randomly sampled opponents for each side;}
    \item \textbf{Random}: baseline which continues to use EvoGym's robot sampling function after the initialization phase.
\end{itemize}

\subsection{Results}


\begin{figure}[ht]
    \centering
    \includegraphics[width=\linewidth]{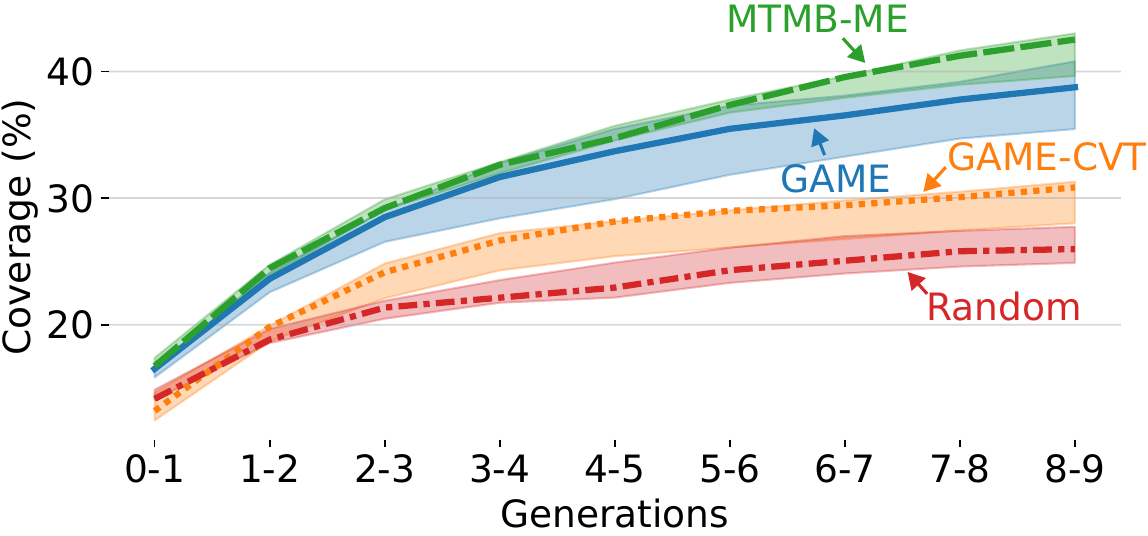}
    \caption{\textbf{Wrestling visual diversity through generations}. \new{ \textbf{GAME} leads to higher coverage than \textbf{GAME-CVT} and \textbf{Random}, but lower than \textbf{MTMB-ME}. The solid line is the median, and the shaded area shows the range from the minimum to the maximum across three replications. }}
    \label{fig:wrestling_coverage}

\end{figure}




\begin{figure}[ht]
    \centering
    \includegraphics[width=\linewidth]{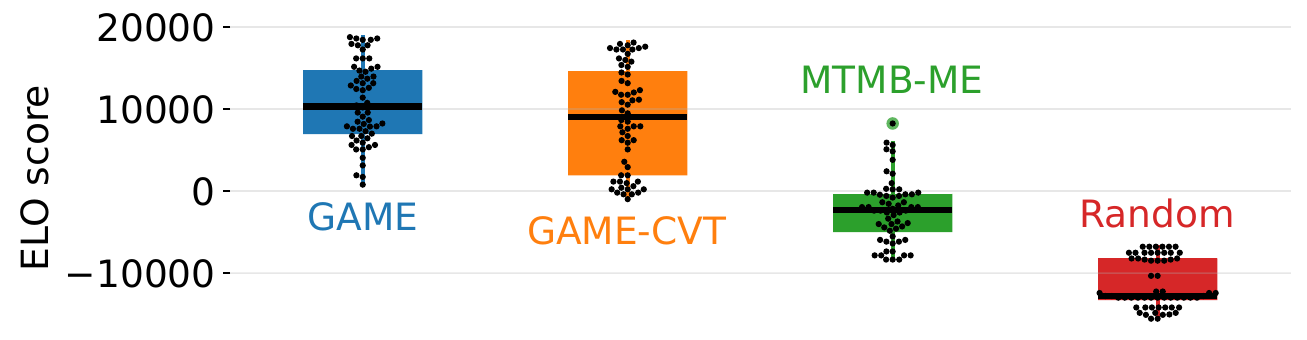}
    \vspace{-0.8cm}
    \caption{\textbf{Wrestling best morphologies' quality}. ELO score from a tournament between the $10$ best morphologies of each side of each variant's replications. \new{\textbf{GAME} and \textbf{GAME-CVT} finds better morphologies than \textbf{MTMB-ME} and \textbf{Random} (p-value $<0.001$ from a Mann–Whitney U test).}}
    \label{fig:wrestling_ELO}
    \vspace{-0.3cm}
\end{figure}

\subsubsection{Main comparison}


\new{\textbf{GAME} leads to a higher coverage with a median of 38.8\% [40.8, 35.5] against 30.8\% for \textbf{GAME-CVT} [31.3, 28.0] and 26.0\% [27.7, 24.9] for \textbf{Random} but lower than for \textbf{MTMB-ME} with 42.5\% [43.0, 39.6] (Fig.~\ref{fig:wrestling_coverage}). Both \textbf{GAME} and \textbf{GAME-CVT} have similar quality when comparing the ELO score of the top-10 elites of each replication in a tournament, both significantly higher than \textbf{MTMB-ME} and \textbf{Random} with a p-value $<0.001$ from a Mann–Whitney U test (Fig.~\ref{fig:wrestling_ELO}).} 

\begin{figure}[ht]
    \centering
    \includegraphics[width=\linewidth]{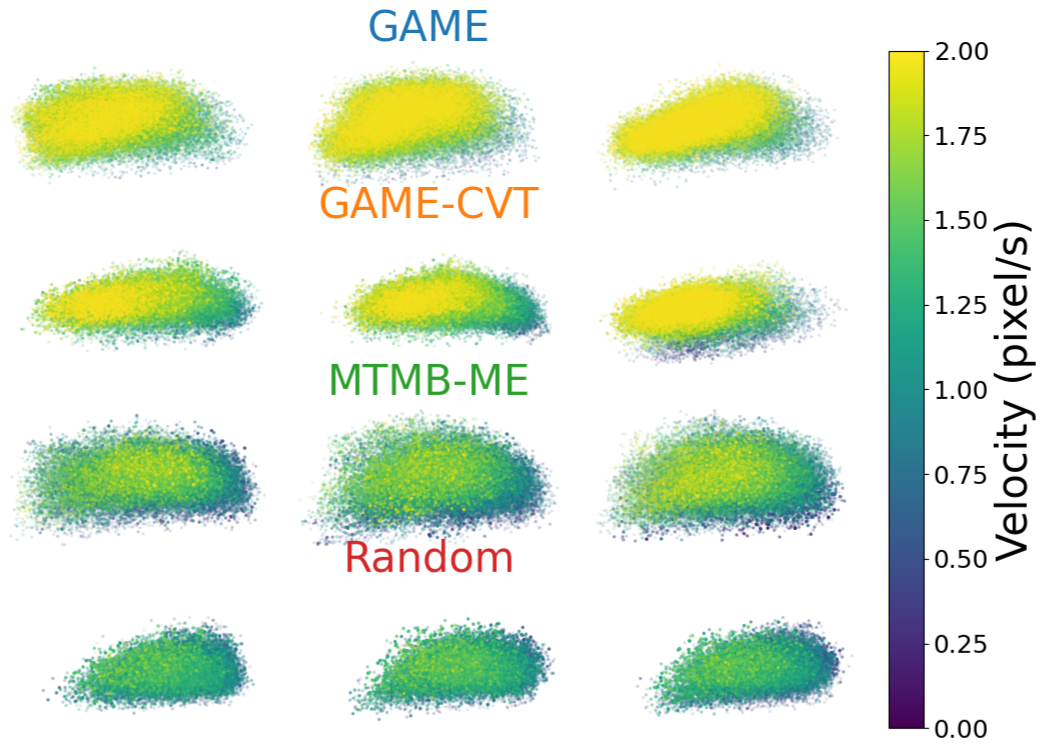}
    \vspace{-0.7cm}
    \caption{\textbf{Wrestling 2D PCA of the visual embedding with velocity}. A single PCA is used for all replications of all variants. \new{\textbf{GAME} and \textbf{GAME-CVT} discover morphologies with higher velocities, which are present more on the left side, suggesting that the VEM embedding PCA captures velocity.}}
    \label{fig:wrestling_PCA_speed}
\end{figure}

\new{To investigate the reason for the higher quality in the tournament, Fig.~\ref{fig:wrestling_PCA_speed} shows the PCA projection (capturing 11.8\% and 7.7\% of the variance) with the average velocity of the two soft robots instead of the fitness (because Wrestling's relative fitness function does not allow us to compare the quality of the solutions) \textbf{GAME} and \textbf{GAME-CVT} leads to finding morphologies with higher velocities than \textbf{MTMB-Me} and \textbf{Random}.} The main component of the PCA of the visual embedding captures slower robots on the right side, which can be visualized in Fig.~\ref{fig:example_morphologies_pca} (and the supplementary video), with faster robots on the left side and slower-moving robots on the right side that do not reach the center of the arena.

\subsubsection{Lineage visualization}
\begin{figure}[ht]
    \centering
    \includegraphics[width=\linewidth]{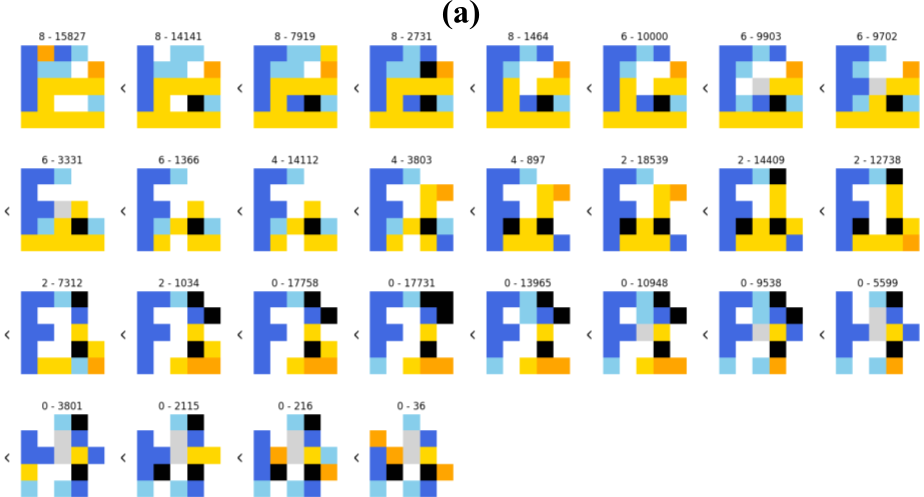}
    \includegraphics[width=\linewidth]{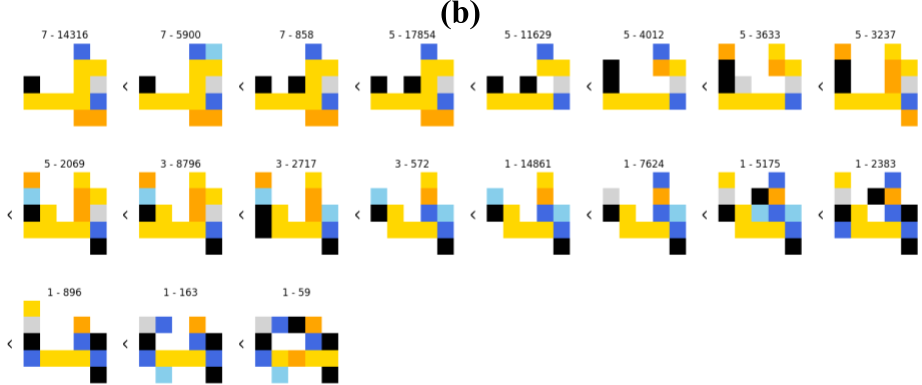}
    \vspace{-0.6cm}
    \caption{\textbf{Ancestry of the best morphologies in the Wrestling tournament}. GAME tracks each solution's genealogy across generations and evaluations. (a) Red. (b) Blue. The variation operator mutates at most $k=3$ voxels, yielding smooth transitions between morphologies. }
    \label{fig:wrestling_best_ancestry}
\end{figure}

GAME keeps track of each ancestor of a solution. Fig.~\ref{fig:wrestling_best_ancestry} shows the full lineage of the best Red and Blue solutions of the inter-variant tournament. Because we mutate at most $k=3$, the solutions change smoothly. In comparison, using a higher $k$ or a crossover could lead to abrupt changes. Still, the initial and final morphologies share very few voxels. Future work could focus on using an indirect encoding~\cite{cheney2014unshackling} to enable crossover and enhance the variation operator's ability to explore the search space.

\subsubsection{Morphological speciation}

\begin{figure}[tb!]
    \centering
    \includegraphics[width=\linewidth]{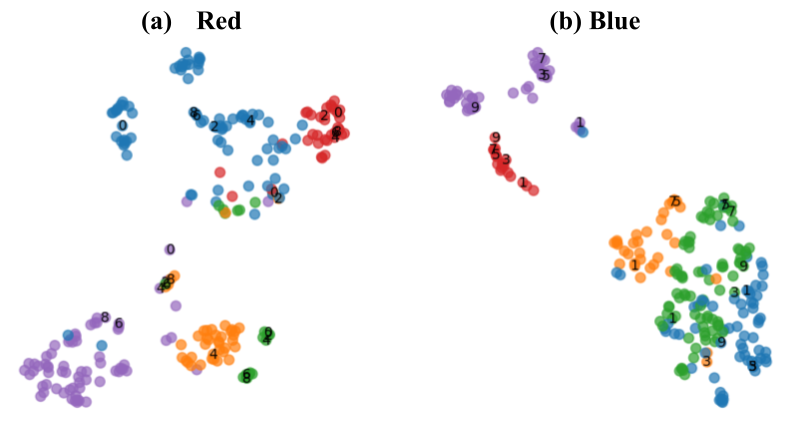}
    \includegraphics[width=\linewidth]{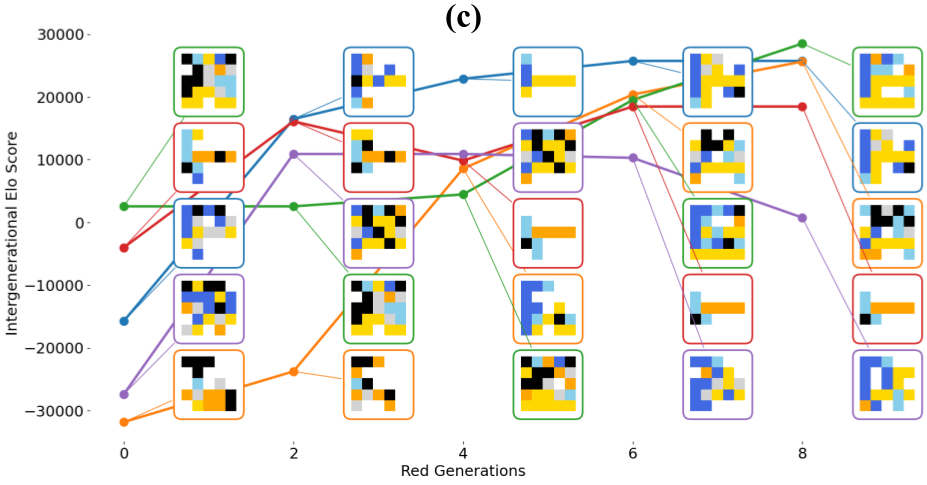}
    \includegraphics[width=\linewidth]{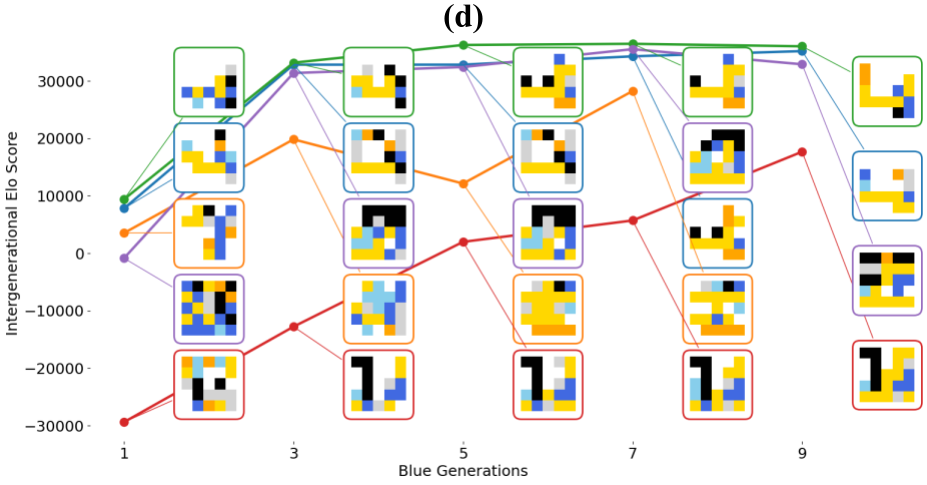}
    \vspace{-0.75cm}
    \caption{\textbf{Morphological species through generations}. (a–b) Morphology clustering into species from all generations using k-modes and UMAP. The numbers indicate the selected elite for each cluster in each generation. (c–d) Average ELO score of each species through generations. GAME discovers all species in the first generation and then improves their average fitness.}
    \label{fig:wrestling_speciation}
    \vspace{-0.5cm}
\end{figure}

To study the evolution of different morphologies across generations, we perform a posteriori morphological speciation analysis on one of \textbf{GAME}'s replications. \new{To perform the clustering, we: (1) aggregate all the elites' morphologies from each generation and (2) use k-modes ($k=5$) to identify distinct morphological clusters that we call species. Compared to k-means, which uses an Euclidean distance, k-modes handles categorical data using the Hamming distance, which is relevant for the morphology defined by voxel types, i.e., $s\in \{0, 1, 2, 3, 4, 5, 6\}^{25}$. To visualize the clustering, we project the morphologies in a 2D space using UMAP, and color each cluster with a different color (Fig.~\ref{fig:wrestling_speciation}.a–b).}
Then, for each cluster at each generation, we compute the average ELO score in an intergenerational tournament. Fig.~\ref{fig:wrestling_speciation}.c–d show the performance over generations for both sides, with snapshots of the best morphology for each cluster and generation.

First, all species are discovered in the first generation, even though clustering is performed a posteriori across all generations. This means that GAME does not discover a significant new morphology after the first generation. This can be explained by two factors: (1) GAME is a QD algorithm that searches for diversity from the start, and (2) this specific problem implementation does not appear open-ended.

Second, for most morphological species, the average ELO score improves over generations, demonstrating GAME's ability to find better solutions. We can observe that one Blue species has disappeared in the last generation. This discrepancy can be attributed to the difference between morphological diversity and the visual embedding diversity optimized by GAME, which captures other elements, such as velocity.

\subsubsection{\new{VEM's robustness to the number of frames}}

\begin{figure}[ht]
    \centering
    \includegraphics[width=\linewidth]{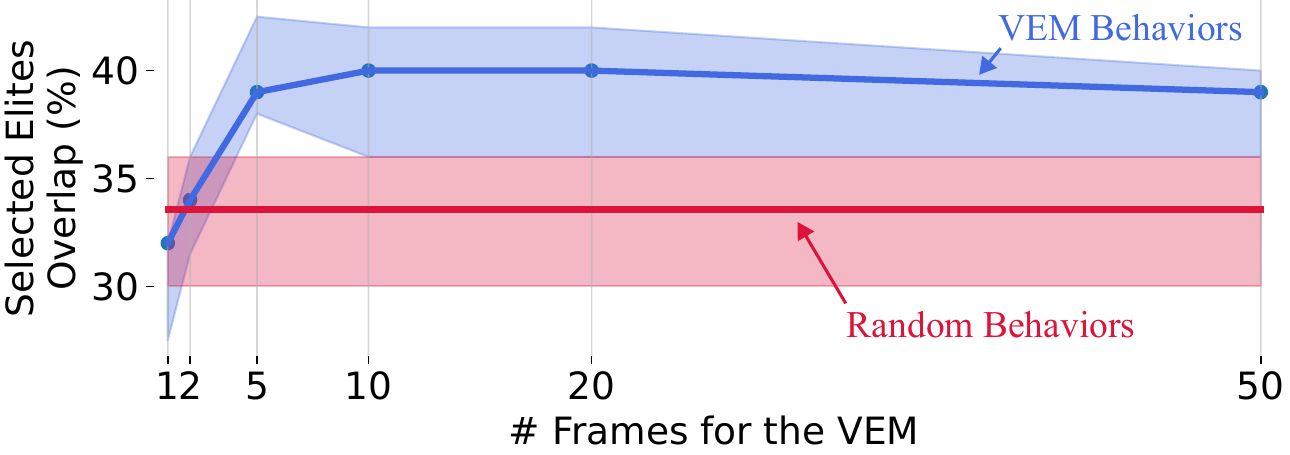}
    \vspace{-0.7cm}
    \caption{\new{\textbf{Robustness to the number of frames used by the VEM}. The VEM behaviors are consistent when at least 5 frames are used and significantly differ from random selection. The solid line is the median, and the shaded area shows the range from the first and third quantiles across 20 replications.}}
    \label{fig:wrestling_selected_elites_overlap}
    \vspace{-0.3cm}
\end{figure}

\new{To evaluate the VEM's robustness to the number of frames used, we picked one final archive from \textbf{GAME} with \num{1000} elites, recomputes the behaviors for different numbers of frames from 1 to 50 (with 5 being the value used in the rest of the paper), and applied the elites selection process (f), and evaluate the overlap of the set of 50 selected elites compared to 50 elites used in the main comparison. As a sanity check, we also evaluate this overlap for randomly sampled behaviors. Because the selection process (\textbf{h}) uses k-means, which is stochastic, we performed 20 replications. Fig~\ref{fig:wrestling_selected_elites_overlap} shows the comparison. The takeaways are: (1) when at least 5 frames are used, the overlap stays constant around 40\%; (2) the overlap is not high, even for 5 frames, due to the stochasticity of k-means, but (3) is still significantly higher than using random behaviors, meaning that the VEM captures some information.}

\subsection{Discussion and Limitations}

\new{\textbf{MTMB-ME} shows higher coverage than \textbf{GAME}. One likely reason is that it does not reset its archives, so it does not waste evaluations finding the same type of behaviors. However, the quality of its elites is significantly lower, supporting the hypothesis that they overfit to random opponents as shown by the lower velocity of its elites in Fig.~\ref{fig:wrestling_PCA_speed}. In the environment, fitness is relative; getting faster against a slow opponent does not increase fitness. One solution could be to use the velocity directly in the fitness function, but this requires expert knowledge that it is part of what is desired. On the other hand, by using adversarial coevolution, \textbf{GAME} promotes higher velocity through an arms race dynamic.}

\new{\textbf{GAME-CVT} having significantly lower coverage validates the use of an unstructured archive. The average number of filled cells per archive is 4.6 (over 20). It is possible that tuning the parameter $N_{cell}$ could yield coverage comparable to or better than that of the unstructured archive. Still, the advantage of the unstructured archive is that it always fills the space, making it much less sensitive to this parameter.}

The main limitation is the use of passive actuation to control the robots, which prevents them from exhibiting reactive behavior based on sensor inputs, a key component of intelligent behavior~\cite {brooks1991intelligence, pfeifer2006body}. 
Future work should focus on integrating body–brain coevolution~\cite{cheney2018scalable, bhatia2021evolution, mertan2023modular, nadizar2025enhancing} into GAME to enable a truly open-ended search space for artificial creatures. \new{For example, \cite{nadizar2025enhancing} show the need for diversity on the brain, body, and behavioral levels to obtain robust creatures.} 

\section{Case Study: Hearthbreaker}

\begin{figure*}[ht]
    \centering
    \includegraphics[width=\linewidth]{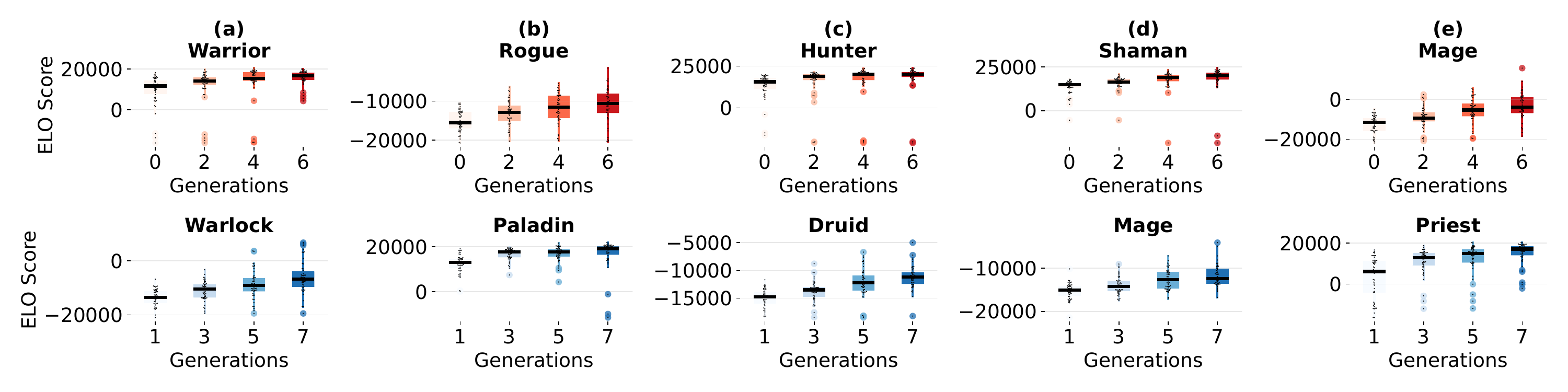}
    \vspace{-0.8cm}
    \caption{\textbf{Hearthbreaker intergenerational tournament}. ELO score of each generation's elites from \textbf{GAME} on the five pairs of classes in Hearthbreaker in an intergenerational tournament. \textbf{GAME} leads to an overall improvement in quality at each generation for both sides and all five pairs of classes.}
    \label{fig:deckbuilding_ELO_generation}
    \vspace{-0.5cm}
\end{figure*}

\begin{figure*}[ht]
    \centering
    \includegraphics[width=\linewidth]{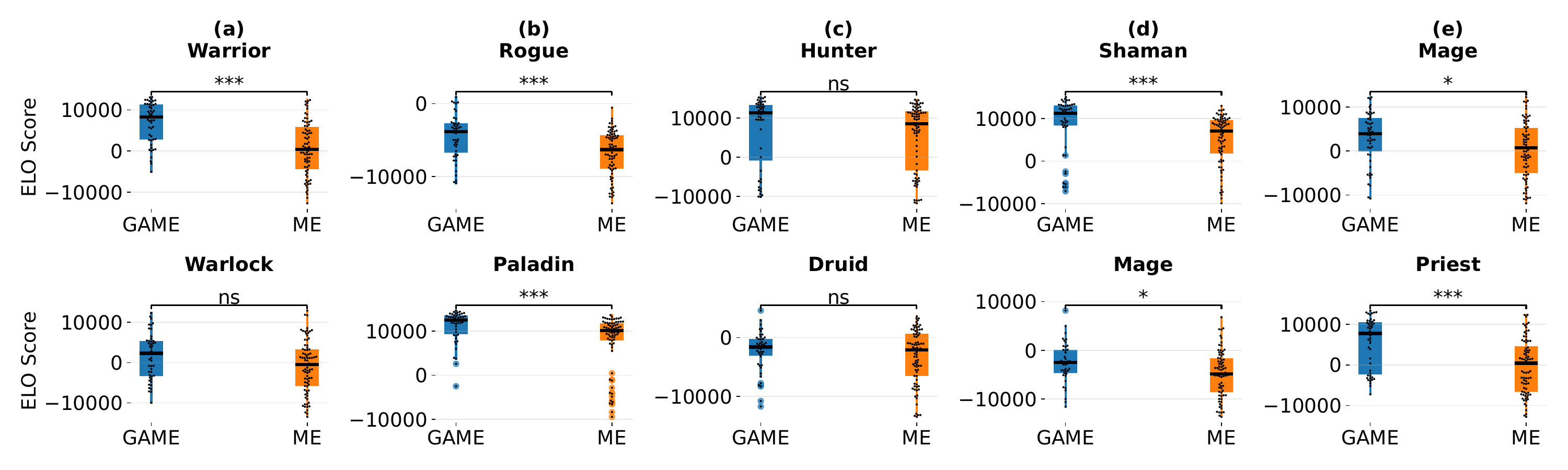}
    \vspace{-0.8cm}
    \caption{\textbf{Hearthbreaker quality comparison between GAME and ME}. \textbf{GAME} leads to better top-50 elites than \textbf{ME} for 7 out of 10 comparisons and is not significantly better for the other 3. (Significance for Mann–Whitney U test: $^{\ast}= p < 0.05$, $^{\ast\ast} = p < 0.01$, $^{\ast\ast\ast} = p < 0.001$, and ns = not significant.)}
    \label{fig:deckbuilding_ELO_comparison}
    \vspace{-0.2cm}
\end{figure*}

\begin{figure*}[ht]
    \centering
    \includegraphics[width=\linewidth]{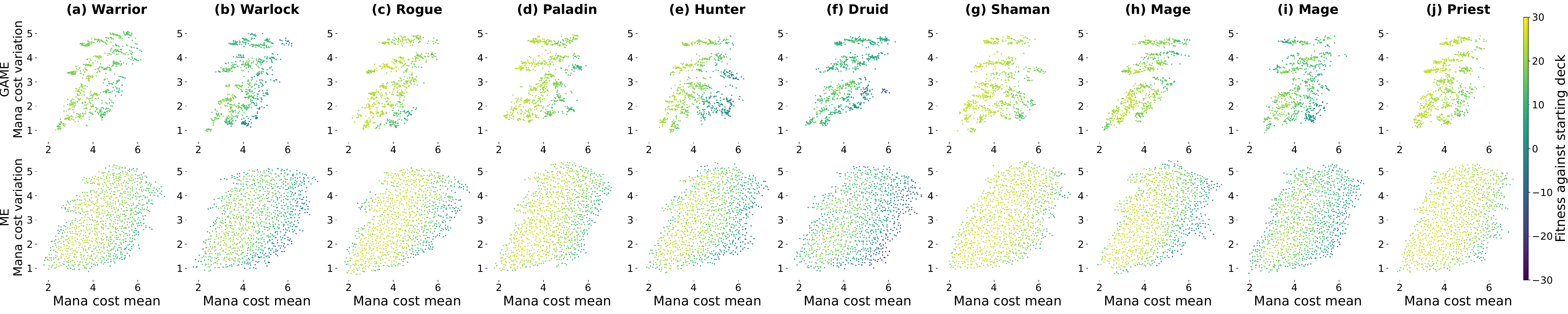}
    \vspace{-0.7cm}
    \caption{\textbf{Hearthbreaker archive comparison between GAME and ME against the starting decks}. The color scale corresponds to the average fitness of the evolved decks against the opponent starting deck for \num{50} duels. \textbf{ME} yields denser, larger coverage by focusing on a single task, whereas \textbf{GAME} illuminates 50 tasks simultaneously. Still, GAME finds solutions with similar quality to \textbf{ME}, even though it was not searching for solutions against the starting decks.}
    \label{fig:deckbuilding_archive_comparison}
    \vspace{-0.3cm}
\end{figure*}

\begin{table*}[ht]
\caption{\textbf{Illumination comparison between GAME and ME in Hearthbreaker against the starting deck.} 
\\ GAME has lower coverage and QD-Score but similar best fitness. }
\centering
\footnotesize
\begin{tabular}{@{}cccccccccccc@{}}
\toprule
\multicolumn{1}{l}{}                 & \multicolumn{1}{l}{\textbf{}} & \multicolumn{2}{c}{\textbf{Warrior VS Warlock}} & \multicolumn{2}{c}{\textbf{Rogue VS Paladin}} & \multicolumn{2}{c}{\textbf{Hunter VS Druid}} & \multicolumn{2}{c}{\textbf{Shaman VS Mage}} & \multicolumn{2}{c}{\textbf{Mage VS Priest}} \\
\multicolumn{1}{l}{\textbf{}}        & \multicolumn{1}{l}{}          & \textbf{(a)}           & \textbf{(b)}           & \textbf{(c)}          & \textbf{(d)}          & \textbf{(e)}          & \textbf{(f)}         & \textbf{(g)}         & \textbf{(h)}         & \textbf{(i)}         & \textbf{(j)}         \\ \midrule
\multirow{2}{*}{\textbf{Coverage}}   & \textbf{GAME}                 & 33.8\%                 & 34.3\%                 & 32.3\%                & 33.5\%                & 35.5\%                & 30.0\%               & 36.3\%               & 27.3\%               & 38.3\%               & 33.0\%               \\
                                     & \textbf{ME}                   & 66.0\%                 & 70.8\%                 & 65.3\%                & 61.5\%                & 66.3\%                & 69.3\%               & 64.0\%               & 63.8\%               & 66.8\%               & 71.5\%               \\
\multirow{2}{*}{\textbf{QD-Score}}   & \textbf{GAME}                 & 6.3                    & 4.7                    & 7.1                   & 7.1                   & 5.6                   & 3.4                  & 8.2                  & 5.8                  & 5.6                  & 7.5                  \\
                                     & \textbf{ME}                   & 12.5                   & 9.1                    & 13.3                  & 13.1                  & 10.8                  & 6.2                  & 14.9                 & 12.1                 & 9.3                  & 16.4                 \\
\multirow{2}{*}{\textbf{Best Elite}} & \textbf{GAME}                 & 24.6                   & 23.7                   & 27.6                  & 27.3                  & 27.1                  & 21.8                 & 26.6                 & 27.5                 & 24.1                 & 27.2                 \\
                                     & \textbf{ME}                   & 26.1                   & 23.4                   & 27.9                  & 27.1                  & 26.7                  & 22.6                 & 28                   & 28.0                 & 25.3                 & 28.2                 \\ \bottomrule
\end{tabular}

\label{tab:deckbuilding_GAME_vs_ME}
\vspace{-0.2cm}
\end{table*}

\subsection{Environment}
\new{To further evaluate GAME's generality, this time without VEM, we apply it to a deck-building and battle card game. In addition, we compare its ability to find diverse and high-quality decks against a one-sided illumination QD ablation, as in \cite{fontaine2019mapping}.} We use a Python simulator of the Hearthstone digital collectible and battle card game~\cite{hearthstone2014}, Hearthbreaker~\cite{hearthbreaker}.

Hearthstone is a two-phase card game. In the first phase, deck building, the player chooses a class (among the $9$ available) and creates a deck of $30$ cards. \new{All classes share a common set of cards (246 cards in the version we used) in addition to the class's unique cards (between 35 and 28) and a unique power usable once each turn. This creates classes with different play styles; for example, the Mage focuses on spells that directly damage a unit, the Warrior on dealing damage with weapons, and the Warlock on flooding its board with cheap minions.} In the second phase, deck battle, the player picks their deck of cards and battles against another player with their own class and deck of cards. The goal is to set the opponent's life from $30$ to $0$. In this case study, similarly to \cite{fontaine2019mapping}, we consider only the deck building phase and use Hearthbreaker's Trade Agent heuristic to play the deck. This heuristic is a greedy policy that maximizes the opponent's minion loss and then maximizes damage to their health.

We followed the method described in \cite{fontaine2019mapping}. The \textbf{solution space} consists of a set of $30$ cards, comprising both class-specific cards and common cards, with no more than two copies of the same card. We used the same \textbf{variation operator}, i.e., GAME picks one elite deck at random, samples $k$ from a geometric distribution with $P(k=1)=0.5$, and then randomly replaces $k$ cards from the elite deck using a uniform distribution over the valid cards. For the \textbf{behavior space}, we did not use the VEM but the same two-dimensional space as in the original paper: the deck's mana cost mean and standard deviation. However, we still used the same unstructured archive because it can handle both small and large spaces with the same hyperparameter, i.e., the number of cells in the archive. We also use the same \textbf{fitness function}: the difference in life between the player and the opponent. Since the game ends when one player reaches $0$, the fitness function is positive when the player wins (and is bounded by $30$) and negative when the player loses (and is bounded by $-30$). Fitness is averaged over $n=50$ duels with fixed seeds to account for stochasticity, with half starting with the evaluated deck and the other half with the opponent's deck.

\subsection{Results}

We compare \textbf{GAME} against an ablation, \textbf{ME}, corresponding to MAP-Elites, i.e., only one generation and one task, with the same total budget of $\num{40000}$ evaluations per side and number of cells in the archive. \textbf{GAME} is set with $N_{gen} = 8$, $N_{budget}=\num{10000}$, $N_{task}=50$, and $N_{cell} = \num{20}$ and \textbf{ME} with $N_{gen} = 1$, $N_{budget}=\num{40000}$, $N_{task}=1$, and $N_{cell} = \num{1000}$. \new{Because each class has unique cards, to simplify the mutation operator, we only evolve decks for one specific class (for each side) on one execution of GAME.} We chose five pairs of classes that are empirically known to be balanced: Warrior and Warlock, Rogue and Paladin, Hunter and Druid, Shaman and Mage, and Mage and Priest. \textbf{ME} independently evolves the deck of one side against the fixed starting deck of the other side (similarly to \cite{fontaine2019mapping}) and is thus executed once for each side. In contrast, \textbf{GAME} coevolves both decks in one run. Fig.~\ref{fig:deckbuilding_ELO_generation} shows the intergenerational ELO score of the five \textbf{GAME} executions on each pair of classes. For all instances, the elite's average quality improves at each generation.

As \textbf{GAME} and \textbf{ME} do not use the same resolution for the archive, for a fairer comparison, we recomputed their archives on a 2D grid with a bin size of $0.1$ and performed a tournament between the new archives' elites for each pair of classes. Fig.~\ref{fig:deckbuilding_ELO_comparison} shows the resulting ELO scores. \textbf{GAME} finds significantly better solutions for 7 out of 10 comparisons and non-significantly better solutions for the remaining three. A simple reason is that \textbf{ME} decks are only evolved to compete against the starting deck, whereas \textbf{GAME} coevolves a diversity of decks that prevent overfitting to a single one.

Table~\ref{tab:deckbuilding_GAME_vs_ME} shows the coverage, QD-Score, and fitness of the best elite found by each method evaluated against the corresponding starting deck.
\textbf{GAME}'s coverage is significantly worse than \textbf{ME}'s. 
The reason is that it splits the $\num{1000}$ cells into $\num{50}$ tasks, thus leaving only $20$ cells for diversity, while \textbf{ME} can allocate the $\num{1000}$ cells for diversity. This becomes apparent in this case study because we use a much smaller 2D behavior space rather than a visual embedding. It is thus much easier to cover the entire space of reachable behaviors. One limitation of \textbf{GAME} is the independence of diversity across tasks, which leads to storing elites with similar behaviors across tasks, thereby reducing the overall diversity of the archive (see Fig.~\ref{fig:deckbuilding_archive_comparison}).  
Still, even though \textbf{GAME}'s elites are not evolved against the starting decks like those from \textbf{ME}, they show similar performance against them. 

\begin{table*}[th]
\footnotesize
\centering
\caption{\new{\textbf{Robustness evaluation of the random seed in Hearthbreaker.} Reevaluation of the intergenerational tournament for one replication of \textbf{GAME} with a new random seed. The mean fitness difference and absolute difference of the \num{40000} evaluations are low, and the rankings of the \num{200} elites are highly correlated (Spearman's rank correlation coefficients have a p-value $<0.001$.)}}
\vspace{-0.2cm}
\label{tab:hearthbreaker_random_seed_robustness}
\begin{tabular}{@{}rccccc@{}}
\toprule
\multicolumn{1}{l}{}                                                                    & \textbf{\begin{tabular}[c]{@{}c@{}}Warrior\\ vs\\ Warlock\end{tabular}} & \textbf{\begin{tabular}[c]{@{}c@{}}Rogue\\ vs\\ Paladin\end{tabular}} & \textbf{\begin{tabular}[c]{@{}c@{}}Hunter \\ vs\\ Druid\end{tabular}} & \textbf{\begin{tabular}[c]{@{}c@{}}Shaman\\ vs\\ Mage\end{tabular}} & \textbf{\begin{tabular}[c]{@{}c@{}}Mage\\ vs\\ Priest\end{tabular}} \\ \midrule
\textbf{Mean fitness difference}          & $-6.4 \times 10^{-4}$ & $4.6 \times 10^{-3}$ & $-1.2 \times 10^{-3}$ & $2.2 \times 10^{-3}$ & $2.0 \times 10^{-3}$  \\
\textbf{Mean fitness absolute difference} & $5.5 \times 10^{-2}$ & $5.7 \times 10^{-2}$ & $5.3 \times 10^{-2}$ & $5.7 \times 10^{-2}$ & $5.7 \times 10^{-2}$ \\
\textbf{Red ELO score Spearman's rank correlation coefficient}  & 0.993                                                                   & 0.993                                                                 & 0.995                                                                 & 0.995                                                               & 0.993                                                               \\
\textbf{Blue ELO score Spearman's rank correlation coefficient} & 0.994                                                                   & 0.995                                                                 & 0.988                                                                 & 0.987                                                               & 0.995                                                               \\ \bottomrule
\end{tabular}
\vspace{-0.3cm}
\end{table*}

\new{To evaluate GAME's robustness to the random seed in Hearthbreaker, we reevaluate the intergenerational tournament for one replication of GAME on all the scenarios with a new random seed (Tab.~\ref{tab:hearthbreaker_random_seed_robustness}). Both the mean fitness differences and the mean absolute differences are low, and the resulting ELO score ranking is significantly correlated with the original ranking (Spearman's rank correlation coefficients are all greater than 0.987, with p-values $<0.001$), validating that GAME finds solutions robust to the stochasticity of this environment. This lower difference relative to Parabellum is due to the fitness having already been averaged over 50 duels.}

\subsection{Limitations}

The current approach has three limitations. First, the behavior space may be too simple and is independent of the duels (as it is purely computed from the deck and not influenced by how the deck is played), which limits the possible diversity of decks. Second, Hearthbreaker's Trade Agent is a greedy heuristic that ignores the combo aspect of the game and focuses only on building decks with many minions (all decks evolved with GAME have at least 86\% (26 out of 30) minion cards, with a median of 97\% (29 out of 30). In contrast, minions only correspond to 65\% of the available cards, which would correspond to 20 out of 30 cards in the deck. Future work should focus on using a more advanced agent, e.g., MCTS~\cite{santos2017monte}, to explore the game's combo aspect and increase the open-endedness of this adversarial problem. \new{Third, we only evolved decks for one class at a time, even though they share 87\% of common cards. Future work could focus on evolving decks for all classes simultaneously to improve sample efficiency, requiring the design of a variation operator that handles the constraints of each class's unique cards.}


\section{Conclusion}
We present GAME, a coevolutionary QD algorithm for illuminating adversarial problems. Combined with a vision-embedding model serving as a domain-agnostic behavior space, GAME requires only videos rather than handcrafted behavior descriptors. We demonstrate GAME's ability to illuminate three different adversarial problems and validate all its components through ablation studies in a multi-agent battle game. We demonstrate its generality by using the same code to evolve both behavior trees for a multi-agent battle game and soft-robot morphologies in a wrestling environment. \new{We also show that GAME finds better solutions than a one-sided illumination baseline on the soft-robot and deck building problems.} Those studies also reveal that GAME is currently limited by the non-open-ended nature of the different search spaces, which prevents it from generating open-ended illumination. Future work should focus on applying GAME to open-ended search spaces, paving the way toward a better understanding of the emergence of open-ended adversarial coevolution.

\section{Acknowledgments} 
Funded by the armasuisse S+T project F00-007.

\footnotesize
\bibliographystyle{IEEEtran}
\bibliography{biblio}

\vfill

\end{document}